# Algorithms and Limits for Compact Plan Representations

**Christer Bäckström**                                 CHRISTER.BACKSTROM@LIU.SE
**Peter Jonsson**                                        PETER.JONSSON@LIU.SE
*Department of Computer Science*
*Linköping University*
*SE-581 83 Linköping, Sweden*

## Abstract

Compact representations of objects is a common concept in computer science. Automated planning can be viewed as a case of this concept: a planning instance is a compact implicit representation of a graph and the problem is to find a path (a plan) in this graph. While the graphs themselves are represented compactly as planning instances, the paths are usually represented explicitly as sequences of actions. Some cases are known where the plans always have compact representations, for example, using macros. We show that these results do not extend to the general case, by proving a number of bounds for compact representations of plans under various criteria, like efficient sequential or random access of actions. In addition to this, we show that our results have consequences for what can be gained from reformulating planning into some other problem. As a contrast to this we also prove a number of positive results, demonstrating restricted cases where plans do have useful compact representations, as well as proving that macro plans have favourable access properties. Our results are finally discussed in relation to other relevant contexts.

## 1. Introduction

The usage and study of representations of objects that are much smaller than the objects themselves is commonplace in computer science. Most of us encounter such representations on a daily basis in the form of zipped files, mp3 files etc. For such practical cases, we usually talk about compressed objects, while the terms compact and succinct are more common in theoretical studies. The meaning of the terms vary but a common and interesting case is when the size of the representation is at most polylogarithmic in the size of the object. Sometimes it is sufficient to just compute a compact representation of an object, for instance, when archiving a file. In other cases the representation must also support various operations efficiently without first unpacking the object into an explicit representation. Performing operations on a compact representation is often harder than performing the same operation on an explicit object, but there are cases when a compact representation can make it easier by emphasising some inherent structure in the object.

One archetypical case of using compact representations is automated planning, although it is seldom viewed in that way. A planning instance is an implicit representation of a graph that is typically exponentially larger than its representation, the instance, and where the solutions, the plans, are paths in this graph. Consider, for example, a STRIPS instance with $n$ variables. These variables implicitly define a state space with $2^n$ states and an action with $m$ preconditions define $2^{n-m}$ arcs in the graph. Similarly, we can define instances where the paths are of exponential length too. Although the planning instances themselves are already compact representations, very little attention has been paid to compact representations of the solutions, which are usually represented explicitly. This paper introduces and analyses a number of such compact representations.

If we first turn to computer science in general we find that compact representations of arbitrary strings is an intensively studied field. For example, Charikar et al. (2005) and Rytter (2003) address the problem of approximating the smallest string representation using a compressed grammar. Bille et al. (2011) show that such representations permit efficient access and matching operations, while Jansson, Sadakane, and Sung (2012) demonstrate representations with efficient edit operations. More structured objects than arbitrary strings can potentially have more compact representations. The following are some examples, displaying positive as well as negative results in various areas. Both Galperin and Wigderson (1983) and Wagner (1986) study the complexity of common graph operations when the graphs are implicitly represented as circuits that tell whether two vertices are connected. Balcázar (1996) uses a variant of that approach to study the complexity of search in AI, using a circuit that generates the adjacency list for a vertex. Bulatov and Dalmau (2006) present an efficient algorithm for certain CSP problems that relies on using a compact representation of the set of solutions. Liberatore and Schaerf (2010) study preprocessing in model checking with focus on the size of the preprocessed parts. Cadoli et al. (2000) study various formalisms for knowledge representation and study when problems modelled in one formalism can be transformed into another formalism with at most a polynomially larger representation.

One approach to compact representations in various areas is the use of macros. This concept has been widely used for a long time also in planning, although seldom for the purpose of providing compact representations. An exception is the following case. The 3S class (Jonsson & Bäckström, 1998b) of planning instances has the property that optimal plans can be of exponential length but it is always possible to decide in polynomial time if there is a plan or not. Giménez and Jonsson (2008) showed that plans for the 3S class always have a polynomial-size representation using macros, and that macro plans can even be generated in polynomial time. That is, although the plan may be of exponential length, and thus necessarily take exponential time to generate, it is possible to generate a compact representation of it in polynomial time. Jonsson (2009) later demonstrated similar results for a number of other classes. Although these particular classes of planning instances may still be too restricted to be of much practical use, the principle of compressing the solution using macros is an interesting tool both for planning and plan explanation.

Other approaches to compact plan representation appear only sparingly in the literature. A notable exception is Liberatore (2005a) who studies two concepts for plan representation that have efficient random access and efficient sequential access respectively. Just like macro plans these are examples of representing one long plan compactly. We might also consider representing a large set of plans compactly. For instance, plan recognition may have to simultaneously consider an exponential number of candidate plans (Geib, 2004). Although seldom viewed in that way, also a reactive plan is a representation of a large set of plans, one for each state from which the goal can be reached. It is, however, known that reactive plans cannot be both compact, efficient and correct in the general case (Jonsson, Haslum, & Bäckström, 2000), although these properties are important, for instance, in spaceship applications (Williams & Pandurang Nayak, 1997). POMDPs may similarly be thought of as a probabilistic variant of reactive plans and compactness of representations is important also in this case (Boutilier & Poole, 1996). Yet another case is when the size of the plan is big but the plan is not necessarily long, which can occur for various types of branching plans, as in contingent planning (Bonet & Geffner, 2000). These three different concepts are not isolated from each other. For instance, Bonet (2010) casts contingent planning into the problem of conformant planning, that is, a branching plan is represented as one long non-branching plan, with the branches appearing as subplans. In all these cases, it is interesting to know if the objects in question have compact representations. Although a compact representation can save space, this may be secondary in many cases. A more important aspect is that if an object has a compact representation then the object has some inherent structure that we may exploit also for other purposes. For instance, if we represent a set of many plans then a representation using recursive macros, or similar, can emphasize the differences and similarities between the plans. This make both comparisons and other operations on the plans more efficient. Similarly, in the case of branching plans we might want to exploit a structure that clearly displays both what two branches have in common and where they differ.

The positive results on macro representations (Giménez & Jonsson, 2008; Jonsson, 2009) prompt the obvious question whether long plans can always be compressed using macros (or any other method). We show in this paper that this is unlikely, no matter what type of compact representation we try to use (macro plans, finite automata or whatever). The remainder of the paper is organized as follows. Section 2 introduces basic notation and concepts as well as the planning framework used in the paper, and it also contains some useful definitions for the complexity results. We then first ask, in Section 3, whether all (optimal) plans for an instance can have compact representations. We find that the answer is no; it is not possible, neither by macros nor any other method. However, the results do not exclude that some plans for each instance can have compact solutions. In Section 4 we thus restrict the question to whether there is a uniform compact representation of one plan for each solvable instance. More precisely, we ask if there is an algorithm that corresponds to one compact representation for each solvable instance. We show that such an algorithm is unlikely to exist if it must also be able to access the actions of the plan in some useful way. In Section 5 we turn to the non-uniform case, asking if each solvable instance has at least one plan that has a compact representation. We primarily consider representations that can efficiently access the actions of the plan sequentially or randomly. We show that also this seems unlikely in the general case, but that there are interesting special cases where such representations do exist. In this section we also investigate macro representations and extend the results by Giménez and Jonsson in two ways. We prove that all plans that have a polynomial-size macro representation can be random accessed in polynomial time without having access to the full plan. However, we also prove that we cannot always represent plans compactly using macros. In Section 6 we analyse whether we can get around the problem of long plans by reformulating planning to some other problem. Also this is answered negatively. If we actually ask for a plan for the original problem, then the problem is inherently intractable also when using reformulation. However, even if considering only the decision problem it still seems not possible to make planning simpler by reformulation. Finally, Section 7 contains a discussion of how the results in the paper are related to and relevant for various other topics like adding information to guide planners, causal graphs and plan explanation. The paper ends with a summary of the results together with a list of open questions.

Some of the results in this paper have appeared in a previous conference publication (Bäckström & Jonsson, 2011b).

## 2. Preliminaries

This section consists of three parts. The first part introduces some general notation and terminology used in the paper. The second part defines the two planning frameworks used in the paper, Finite Functional Planning and propositional STRIPS, and presents some constructions that will be frequently used. The third part briefly recapitulates the concept of advice-taking Turing machines and also defines the 3SAT problem that will be used on several occasions in the paper.

### 2.1 General Notation and Terminology

A sequence of objects $x_1, x_2, \ldots, x_n$ is written $\langle x_1, x_2, \ldots, x_n \rangle$, with $\langle \rangle$ denoting the empty sequence. Given a set $X$ of objects, the set of all sequences over $X$, including $\langle \rangle$, is denoted $X^*$. For a set, sequence or other aggregation $X$ of objects, we write $|X|$ to denote the cardinality (the number of objects) of $x$ and we write $||X||$ to denote the size (the number of bits of the representation) of $x$. The composition of two functions $f$ and $g$ is denoted $f \circ g$ and is defined as $(f \circ g)(x) = f(g(x))$.

The *negation* of a propositional atom $x$ is denoted $\overline{x}$. A *literal* is either an atom or its negation and the set $L(X)$ of literals over a set $X$ of atoms is defined as $L(X) = \{x, \overline{x} \mid x \in X\}$. Negation is extended to literals such that $\overline{\overline{\ell}}$ is the same literal as $\ell$. Negation is also extended to sets such that if $X$ is a set of literals then $\overline{X} = \{\overline{\ell} \mid \ell \in X\}$. Let $Y$ be a subset of $L(X)$ for some set $X$ of atoms. Then $Pos(Y) = \{x \in X \mid x \in Y\}$ is the set of atoms that appear positive in $Y$, $Neg(Y) = \{x \in X \mid \overline{x} \in Y\}$ is the set of atoms that appear negated in $Y$ and $Atoms(Y) = Pos(Y) \cup Neg(Y)$. The set $Y$ is *consistent* if $Pos(Y) \cap Neg(Y)$ is empty and a set $Z$ of atoms *satisfies* $Y$ if both $Pos(Y) \subseteq Z$ and $Neg(Y) \cap Z = \varnothing$. The *update operator* $\ltimes$ is a binary function such that given a set $X$ of atoms and a set $Y$ of literals, $X \ltimes Y$ is a set of atoms defined as $X \ltimes Y = (X - Neg(Y)) \cup Pos(Y)$.

### 2.2 Planning

For positive results on compact representations, we want the results to apply to as general and powerful planning languages as possible, so the results hold also for all languages that are more restricted. Hence, we use the Finite Functional Planning formalism (Bäckström & Jonsson, 2011a), which makes a minimum of assumption about the language, except that it is a ground language over state variables with finite domains.

**Definition 1.** A *Finite Functional Planning (FFP) frame* is a tuple $\langle V, D, A \rangle$ where $V$ is an implicitly ordered set of *variables*, $D : V \to \mathbb{N}$ is a *domain function* that maps every variable to a finite subset of the natural numbers and $A$ is a set of *actions*. The frame implicitly defines the *state space* $S(\boldsymbol{f}) = D(v_1) \times \ldots \times D(v_n)$, where $v_1, \ldots, v_n$ are the variables in $V$ in order. The members of $S(\boldsymbol{f})$ are referred to as *states*. Each action $a$ in $A$ has two associated total functions, the *precondition* $\mathrm{pre}(a) : S(\boldsymbol{f}) \to \{0, 1\}$ and the *postcondition* $post(a) : S(\boldsymbol{f}) \to S(\boldsymbol{f})$. For all pairs of states $s, t \in S(\boldsymbol{f})$ and actions $a \in A$, $a$ is *from $s$ to $t$* if both

    1) $pre(a)(s) = 1$ and

    2) $t = post(a)(s)$.

A sequence $\omega = \langle a_1, \ldots, a_\ell \rangle \in A^*$ is a *plan* from a state $s_0 \in S(\boldsymbol{f})$ to a state $s_\ell \in S(\boldsymbol{f})$ if either

    1) $\omega = \langle \rangle$ and $s_0 = s_\ell$ or

    2) there are states $s_1, \ldots, s_{\ell-1} \in S(\boldsymbol{f})$ such that $a_i$ is from $s_{i-1}$ to $s_i$ (for $1 \leq i \leq \ell$).

An *FFP instance* is a tuple $\boldsymbol{p} = \langle V, D, A, I, G \rangle$ where $\boldsymbol{f} = \langle V, D, A \rangle$ is an FFP frame, $I \in S(\boldsymbol{f})$ is a state and $G : S(\boldsymbol{f}) \to \{0, 1\}$ is a total function. A state $s \in S(\boldsymbol{f})$ is a *goal state* for $\boldsymbol{p}$ if $G(s) = 1$. The goal $G$ is *reachable* from a state $s \in S(\boldsymbol{f})$ if there is a plan from $s$ to some goal state for $\boldsymbol{p}$. A solution for $\boldsymbol{p}$ is a plan from $I$ to goal state $s \in S(\boldsymbol{f})$. A solution for $\boldsymbol{p}$ is called a *plan* for $\boldsymbol{p}$.

The complexity of computing the pre- and postconditions of the actions and of the goal function is referred to as *step complexity*. In this paper, we will only consider the subclass FFP([**P**]) which consists of all FFP frames and instances with polynomial step complexity. We will occasionally also consider restrictions of FFP([**P**]) and use the notation FFP($p$) for the class of all FFP frames $\boldsymbol{f}$ (and instances $\boldsymbol{p}$) where the action pre- and postconditions (and $G$) can all be computed in $p(||\boldsymbol{f}||)$ time (and in $p(||\boldsymbol{p}||)$ time), where $p$ is a polynomial. We furthermore say that an FFP([**P**]) instance $\boldsymbol{p} = \langle V, D, A, I, G \rangle$ is *deterministic* if for all $s \in S(\boldsymbol{p})$ such that $\boldsymbol{p}$ has a plan from $I$ to $s$, there is at most one $a \in A$ such that $pre(a)(s) = 1$. That is, if an instance is deterministic then the planner is never faced with a choice between two or more actions.

When proving that no compact representation can exist, the result gets stronger if we use a weaker formalism. That is, we want to use the most restricted formalism possible, since the results will then automatically apply to all formalisms that are more expressive. Hence, we will use propositional STRIPS for such results. There are a number of common variants of propositional STRIPS that are known to be equivalent to each other and to the SAS$^+$ formalism under a strong form of polynomial reduction (Bäckström, 1995). What we will refer to as STRIPS in this paper is the variant called *propositional STRIPS with negative goals (PSN)* by Bäckström. It can be defined as a special case of FFP([**P**]) that uses binary variables, but we define it here in a more traditional way, treating the variables as propositional atoms.

**Definition 2.** A STRIPS *frame* is a tuple $\boldsymbol{f} = \langle V, A \rangle$ where $V$ is a set of propositional atoms and $A$ is a set of actions. The state space is defined as $S(\boldsymbol{f}) = 2^V$ and states are subsets of $V$. Each action $a$ in $A$ has a precondition $pre(a)$ and a postcondition $post(a)$, which are both consistent sets of literals over $V$. For all pairs of states $s, t \in S(\boldsymbol{f})$ and actions $a \in A$, $a$ is *from $s$ to $t$* if both

    1) $s$ satisfies $pre(a)$ and

    2) $t = s \ltimes post(a)$.

A sequence $\omega = \langle a_1, \ldots, a_\ell \rangle \in A^*$ is a *plan* from a state $s_0 \in S(\boldsymbol{f})$ to a state $s_\ell \in S(\boldsymbol{f})$ if either

    1) $\omega = \langle \rangle$ and $s_0 = s_\ell$ or

    2) there are states $s_1, \ldots, s_{\ell-1} \in S(\boldsymbol{f})$ such that $a_i$ is from $s_{i-1}$ to $s_i$ (for $1 \leq i \leq \ell$).

A STRIPS *instance* is a tuple $\boldsymbol{p} = \langle V, A, I, G \rangle$ such that $\boldsymbol{f} = \langle V, A \rangle$ is a STRIPS frame, $I$ is a state in $S(\boldsymbol{f})$ and $G$ is a consistent set of literals over $V$. A state $s \in S(\boldsymbol{f})$ is a *goal state* for $\boldsymbol{p}$ if $s$ satisfies $G$. The goal $G$ is *reachable* from a state $s \in S(\boldsymbol{f})$ if there is a plan from $s$ to some goal state for $\boldsymbol{p}$. A solution for $\boldsymbol{p}$ is a plan from $I$ to some goal state $s \in S(\boldsymbol{f})$. A solution for $\boldsymbol{p}$ is called a *plan* for $\boldsymbol{p}$.

The notation $a \colon X \Rightarrow Y$ will be frequently used to define an action $a$ with precondition $X$ and postcondition $Y$.

All negative results will be proven to hold for STRIPS. However, in most cases the results hold even for many restricted subclasses of STRIPS. It would lead too far to survey such cases in this paper so we will use the restriction to unary actions as an archetypical case throughout the paper.

**Definition 3.** A STRIPS action $a$ is *unary* if $|\text{post}(a)| = 1$, a set of STRIPS actions is unary if all its actions are unary and a STRIPS frame or instance is unary if its action set is unary.

Unary actions may seem like a very limiting restriction but has been demonstrated as sufficient in many cases for use in on-board controllers for spacecrafts (Muscettola et al., 1998; Brafman & Domshlak, 2003). This is not surprising, though, since STRIPS planning is **PSPACE**-complete and remains so even when restricted to unary actions (Bylander, 1994). Given a STRIPS instance it is always possible to construct a corresponding STRIPS instance that is unary. The following reduction to unary instances is a simplified STRIPS version of the reduction used for $SAS^+$ (Bäckström, 1992, proof of Theorem 6.7).

**Construction 4.** Let $\boldsymbol{p} = \langle V, A, I, G \rangle$ be a STRIPS instance. Construct a corresponding instance $\boldsymbol{p}' = \langle V', A', I', G' \rangle$ as follows. Define $V_{lock} = \{v_{lock}^a \mid a \in A\}$. Then let $V' = V \cup V_{lock}$, $I' = I$ and $G' = G \cup \overline{V_{lock}}$. Define $A'$ such that for each $a \in A$, it contains the following actions:

$a_{begin} \colon \text{pre}(a) \cup \overline{V_{lock}} \Rightarrow \{v_{lock}^a\}$,
$a_{end} \colon \text{post}(a) \cup \{v_{lock}^a\} \Rightarrow \{\overline{v_{lock}^a}\}$,
$a_i \colon \{v_{lock}^a\} \Rightarrow \{\ell_i\}$, for each $\ell_i \in \text{post}(a)$.

We leave it without proof that this construction is a polynomial reduction from the class of STRIPS instances to the class of unary STRIPS instances. It is furthermore worth noting that the construction can easily be modified to use padding with redundant variables to make all original actions correspond to the same number of actions in the unary instance. Hence, it is possible to make a reduction where the plans for the unary instance are at most a constant factor longer than the corresponding plans for the original instance.

We will also make frequent use of STRIPS instances that include encodings of binary counters based on the following construction, which uses one action for each bit and can increment a non-negative integer encoded in binary.

**Construction 5.** An $n$-bit binary counter can be encoded in STRIPS as follows: let $V = \{x_1, \ldots, x_n\}$ and let $A$ contain the $n$ actions

$$a_i \colon \{\overline{x_i}, x_{i-1}, \ldots, x_1\} \Rightarrow \{x_i, \overline{x_{i-1}}, \ldots, \overline{x_1}\} \quad (1 \le i \le n).$$

The following is a plan for counting from 0 to 16 using a 5-bit counter according to Construction 5:

$$\langle a_1, a_2, a_1, a_3, a_1, a_2, a_1, a_4, a_1, a_2, a_1, a_3, a_1, a_2, a_1, a_5 \rangle.$$

While we could modify the binary counter to use only unary actions as described in Construction 4, a more direct way to get unary actions is to count in Gray code.

**Construction 6.** (Bäckström & Klein, 1991) An $n$-bit Gray-code counter can be encoded in Strips as follows: let $V = \{x_1, \ldots, x_n\}$ and let $A$ contain the $2n$ actions

$$s_i \colon \{\overline{x_i}, x_{i-1}, \overline{x_{i-2}}, \ldots, \overline{x_1}\} \Rightarrow \{x_i\} \quad (1 \leq i \leq n),$$
$$r_i \colon \{x_i, x_{i-1}, \overline{x_{i-2}}, \ldots, \overline{x_1}\} \Rightarrow \{\overline{x_i}\} \quad (1 \leq i \leq n).$$

The following is a plan for counting from 0 to 16 with a 5-bit Gray-code counter according to Construction 6:

$$\langle s_1, s_2, r_1, s_3, s_1, r_2, r_1, s_4, s_1, s_2, r_1, r_3, s_1, r_2, r_1, s_5 \rangle.$$

## 2.3 Complexity Theory

We will use the abbreviation DTM for *deterministic Turing machine* and NTM for *nondeterministic Turing machine*. In addition to these standard types, we will also use advice taking Turing machines of both deterministic and nondeterministic type.

An *advice-taking Turing machine* $M$ has an associated sequence $a_1, a_2, a_3, \ldots$ of *advice strings*, a special *advice tape* and an *advice function* $a$, from the natural numbers to the advice sequence, such that $a(n) = a_n$. On input $x$ the advice tape is immediately loaded with $a(||x||)$. After that $M$ continues in the normal way, except that it also has access to the advice written on the advice tape. If there exists a polynomial $p$ such that $||a(n)|| \leq p(n)$, for all $n > 0$, then $M$ is said to use *polynomial advice*. The complexity class **P/poly** is the set of all decision problems that can be solved by some advice-taking DTM that runs in polynomial time using polynomial advice. This can be extended such that, for instance, **NP/poly** is defined by the NTMs that run in polynomial time using polynomial advice. Note that the advice depends only on the size of the input, not on its content. Furthermore, the advice sequence must only exist; it does not need to be computable. The following two results from the literature will be used later in this paper.

**Theorem 7.** *a) If* $\boldsymbol{NP} \subseteq \boldsymbol{P/poly}$, *then the polynomial hierarchy collapses (Karp & Lipton, 1980, Theorem 6.1). b) Let* $k > 0$ *be an integer. If* $\boldsymbol{\Pi}_k^p \subseteq \boldsymbol{\Sigma}_k^p / \boldsymbol{poly}$, *then the polynomial hierarchy collapses to level* $k + 2$ *(Yap, 1983, Lemma 7 combined with Theorem 2).*

The *3SAT problem* consists of instances of the form $C = \{c_1, \ldots, c_m\}$ where each $c_i$, for $1 \leq i \leq m$, is called a *clause* and is a set of exactly three literals over some universe of binary variables. The instance $C$ is *satisfiable* if there exists some assignment of truth values to the variables used in $C$ such that at least one literal is true in each $c_i \in C$. It is otherwise *unsatisfiable*. Deciding satisfiability for 3SAT is **NP**-complete, while deciding unsatisfiability is co**NP**-complete. More precisely, we will use the following definition of 3SAT in this paper.

**Definition 8.** For all integers $n > 0$, let $X_n = \{x_1, \ldots, x_n\}$ be a set of variables and let $m(n)$ be the number of possible 3-literal clauses over $X_n$. Let $c_n^1, c_n^2, \ldots, c_n^{m(n)}$ be some fixed systematic enumeration of these clauses and let $C_n = \{c_n^1, c_n^2, \ldots, c_n^{m(n)}\}$. Each clause $c_n^i$ defines three literals such that[1] $c_n^i = \{\ell_i^1, \ell_i^2, \ell_i^3\}$. Further, let $C_n^0, C_n^1, \ldots, C_n^{2^{m(n)}-1}$ be a fixed systematic enumeration of all subsets of $C_n$, and let $\boldsymbol{s}_n^i = \langle X_n, C_n^i \rangle$, for $0 \leq i < 2^{m(n)}$. Also implicitly define the set $E_n = \{e_n^1, e_n^2, \ldots, e_n^{m(n)}\}$ of atoms and its subsets $E_n^i = \{e_n^j \mid c_n^j \in C_n^i\}$, for $0 \leq i < 2^{m(n)}$.

The sequence $\boldsymbol{s}_n^0, \boldsymbol{s}_n^1, \ldots, \boldsymbol{s}_n^{2^{m(n)}-1}$ is a systematic enumeration of all possible **3SAT** instances over $n$ variables, and hence equivalent to the usual definition of **3SAT**. Technically speaking, this is a redundant encoding of **3SAT** since it allows instances that specify more variables than are used in the clauses. This is harmless, however; all non-redundant instances remain, so we still have all hard instances and neither of the redundantly encoded instances can be harder than their non-redundant counterpart. Since $m(n) \leq 8n^3$, the enumerations of $C_n$ and $E_n$ can be chosen such that they are polynomial-time computable, and we assume some such enumerations have been fixed from now on. We also note that a set $E_n^i$ uniquely identifies the clause set $C_n^i$.

## 3. Representing Arbitrary Plans Compactly

It is known that there are cases where planning instances have exponential-size plans but the plans always have a polynomial-size representation (Giménez & Jonsson, 2008). An obvious question is thus whether all plans, including those of exponential length, can have polynomial representations. For one interpretation of the question the answer is trivially yes.

**Observation 9.** *The set of all plans for an arbitrary FFP([$\boldsymbol{P}$]) instance $\boldsymbol{p}$ has an $O(\|\boldsymbol{p}\|)$ size representation, since the instance itself together with a deterministic planning algorithm that successively enumerates and outputs all plans is such a representation.*

Although this is a trivial and not very useful observation it highlights some fundamental issues of representations. Liberatore (2005a) discusses a similar representation, where instead of specifying an algorithm he defines a lexiographic ordering on the actions. Furthermore, he adds a plan index to be able to represent a single plan rather than just the whole set of plans. However, such an index is not unproblematic, as we will see soon.

A more interesting interpretation of our question is whether every single plan for a particular instance can have a polynomial representation. Even more precisely, is there a polynomial $p$ such that every plan for every planning instance has a representation of size $O(p(n))$, where $n$ is the size of the planning instance? To investigate this question we consider the most simple compact notation possible, an index number $i$ for each plan for a particular instance. Since instances may have infinitely many plans, due to cycles in the state-transition graph, we consider optimal plans only. There is, however, no guarantee that even such an index is small enough—a polynomial number of bits may not be sufficient to represent it.

---

[1]. We sometimes omit index $n$, when it can be assumed obvious from context, and thus write $\ell_i^k$ rather than $\ell_{n,i}^k$.

**Construction 10.** Given an arbitrary integer $n > 0$, construct the STRIPS instance $\boldsymbol{p}_n = \langle V_n, A_n, I_n, G_n \rangle$ such that $V_n = \{x_1, \ldots, x_n, y\}$, $I_n = \varnothing$, $G_n = \{x_1, \ldots, x_n\}$ and $A_n$ contains the actions

$$a_i \colon \{\overline{x_i}, x_{i-1}, \ldots, x_1\} \Rightarrow \{x_i, \overline{x_{i-1}}, \ldots, \overline{x_1}, \overline{y}\} \quad (1 \leq i \leq n)$$
$$b_i \colon \{\overline{x_i}, x_{i-1}, \ldots, x_1\} \Rightarrow \{x_i, \overline{x_{i-1}}, \ldots, \overline{x_1}, y\} \quad (1 \leq i \leq n).$$

**Lemma 11.** *For every integer $n > 0$, instance $\boldsymbol{p}_n$ according to Construction 10 has $2^{2^n-1}$ optimal plans.*

*Proof.* Let $n > 0$ be an arbitrary integer and $\boldsymbol{p}_n$ a corresponding STRIPS instance according to Construction 10. This instance is a binary counter over the variables $x_1, \ldots, x_n$ as in Construction 5, except that it has an extra variable $y$ that can be independently set to true or false, depending on whether an action of type $a_i$ or $b_i$ is chosen. Since the variables $x_1, \ldots, x_n$ can be interpreted as a binary number, let the notation $\langle m, \overline{y} \rangle$ represent the state where $x_1, \ldots, x_n$ encodes the number $m$ and $y$ is false, and let $\langle m, y \rangle$ represent the corresponding state where $y$ is true. Whenever in a state $\langle m, \overline{y} \rangle$ or $\langle m, y \rangle$ (where $m < 2^n - 1$) it is possible to go to either of $\langle m+1, \overline{y} \rangle$ or $\langle m+1, y \rangle$ using one action, but to no other states. The state transition graph for this instance appears in Figure 1. The initial state is $\langle 0, \overline{y} \rangle$ and the only goal states are $\langle 2^n - 1, \overline{y} \rangle$ and $\langle 2^n - 1, y \rangle$. Hence, any plan for $\boldsymbol{p}_n$ must be of length $2^n - 1$. From every state that is not a goal state there are two different actions to choose between and they lead to different states. However, the goal is reachable from both these states so the choice of action does not matter. Hence, there are $2^{2^n-1}$ different plans for $\boldsymbol{p}_n$. □

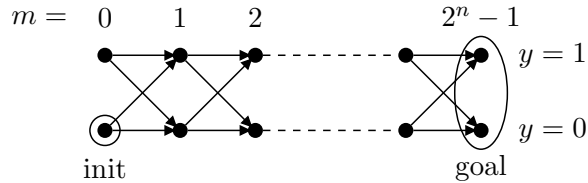

Figure 1: State-transition graph for proof of Lemma 11.

Although the atom $y$ is redundant in this particular example the whole construction could be a part of a larger instance, where $y$ does have a purpose. It should also be noted that the instances used in the proof have only optimal plans; all plans have the same length. All is now set to prove the previous claim.

**Theorem 12.** *For every integer $n > 0$, it takes $2^n - 1$ bits to index all optimal plans for instance $\boldsymbol{p}_n$ according to Construction 10.*

*Proof.* Since an $m$-bit number can distinguish between at most $2^m$ different objects, it follows from Lemma 11 that at least $2^n - 1$ bits is necessary to index the plans for $\boldsymbol{p}_n$ □

This result immediately implies that not all (optimal) plans for a STRIPS instance can have polynomial-size representations. This holds even under some restrictions, like unary actions. If basing Construction 10 on a Gray counter instead of a binary counter then every action has two postconditions. Rewriting this using Construction 4 yields an equivalent instance with unary actions using a block of four actions for each action in the original instance. Although the plans get 4 times longer the number of plans will remain the same. Hence, Theorem 12 still holds.

This theorem leaves the possibility open that some of the plans for an instance can have polynomial representations, although not all of them can. An interesting question is thus how many of the plans for an instance can have polynomial representations? To answer that question we stray into the field of information theory and Kolmogorov complexity. It is out of the scope of this paper to treat that field in detail, but loosely speaking, the Kolmogorov complexity of a string is the size of the smallest DTM that can generate the string with no input. Let $K(x)$ denote the Kolmogorov complexity of a binary string $x$. The following lemma is due to Buhrman et al. (2000, Lm. 1).

**Lemma 13.** *(Incompressibility lemma) Let $c$ be a positive integer. Every set $A$ of cardinality $m$ has at least $m(1 - 2^{-c}) + 1$ elements $x$ with $K(x) \geq \lfloor \log m \rfloor - c$.*

This lemma can be used to show that the fraction of plans that can be compactly represented approaches zero as the size of instances approaches infinity.

**Theorem 14.** *Let $p$ be an arbitrary polynomial. Consider instances $\boldsymbol{p}_n$ according to Construction 10 for arbitrary integers $n > 0$. Let $t(n)$ be the total number of plans for $\boldsymbol{p}_n$ and let $s(n)$ be the number of plans that can be represented with at most $p(n)$ bits. Then $\lim_{n \to \infty} \frac{s(n)}{t(n)} = 0$.*

*Proof.* Let $p$ be an arbitrary polynomial. We know from Lemma 11 that $t(n) = 2^{2^n - 1}$. For every $n > 0$, let $c(n) = 2^n - p(n) - 2$. The incompressibility lemma then says that there are at least $t(n)(1 - 2^{-c(n)}) + 1$ plans $\omega$ such that $K(\omega) \geq \lfloor \log t(n) \rfloor - c(n)$. That is, there are at most $2^{-c(n)}t(n) - 1$ plans $\omega$ such that $K(\omega) < \lfloor \log t(n) \rfloor - c(n)$. Using the values for $t(n)$ and $c(n)$ above, this simplifies to say that there are at most $2^{p(n)+1} - 1$ plans $\omega$ such that $K(\omega) \leq p(n)$. Hence, $s(n) \leq 2^{p(n)+1} - 1$. The theorem then follows since $0 \leq \lim_{n \to \infty} \frac{s(n)}{t(n)} \leq \lim_{n \to \infty} \frac{2^{p(n)+1} - 1}{2^{2^n - 1}} = 0$. □

This means that even if it is the case that some plans for every solvable instance can have compact representations, the probability that a particular plan has a compact representation will be vanishingly low for large instances. Although it is not strictly necessary to use Kolmogorov complexity to prove Theorem 14 doing so makes the information-theoretic aspect of compact representations clearer.

## 4. Uniform Compact Representations of Plans

We now know that we cannot, in general, compress arbitrary exponential plans to subexponential size. But what if we do not choose ourselves which plan to use? The previous result still leaves open for the possibility that a small fraction of solutions for a planning instance could have compact representations. However, the planner (or an oracle or whatever) would then have to choose for us which plan to present us with a compact representation of. Suppose a planner could actually do this, how would we make use of it? If we still need the actual plan itself then we cannot avoid its exponential size. Hence, the interesting case seems to be if we could at least access useful information in the plan efficiently.

The term representation is used in a loose sense here, but need not really be precisely defined for the moment. It suffices to note that any representation needs both some kind of data structure and some kind of access algorithm, with the extreme cases being either a vector of data with the trivial access algorithm or an algorithm that embeds all the data.

What could it mean to access a compact representation efficiently? We will investigate two such criteria. The first one is that we can efficiently retrieve the actions of the actual plan sequentially. Our interpretation of efficient will be that actions can be retrieved with *polynomial delay* (Johnson, Papadimitriou, & Yannakakis, 1988). The second criterion is that any action in the actual represented plan can be random accessed in polynomial time, in the size of the instance.

Before looking at explicit representations for each plan we take a look at the uniform case, where we have a single representation that covers all instances. More precisely, we consider the case of a single algorithm that works as a compact representation for some plan for every solvable instance.

**Theorem 15.** *If there is an algorithm that for any solvable* STRIPS *instance $\boldsymbol{p}$ can either generate some plan for $\boldsymbol{p}$ sequentially with polynomial delay or random access any action in some plan for $\boldsymbol{p}$ in polynomial time, then $\boldsymbol{P} = \boldsymbol{NP}$.*

Note that this theorem does not follow from the fact that STRIPS planning is **PSPACE**-complete since only solvable instances are considered. Before proving the theorem we need to introduce some extra technical machinery. We start by encoding 3SAT instances according to Definition 8 in STRIPS as follows.

**Construction 16.** Let $n > 0$ and $i$, where $0 \leq i < 2^{m(n)}$, be arbitrary integers. Construct the STRIPS instance $\boldsymbol{p}_n^i = \langle V_n, A_n, E_n^i, \{goal\} \rangle$ such that $V_n = X_n \cup E_n \cup \{cts, ctu, goal, inc\} \cup \{v_0, \ldots, v_{m(n)}\}$ and $A_n$ has the actions specified in Table 1.

As previously noted, each subset $E_n^i$ of $E_n$ uniquely identifies the 3SAT instance $\boldsymbol{s}_n^i$ by telling which clauses in $C_n$ are 'enabled' in $\boldsymbol{s}_n^i$. That is, the initial state selects the particular $n$-variable instance we are interested in. The actions are partitioned into three groups. Group I contains the two actions $acs$ and $acu$, which set the atoms $cts$ and $ctu$ respectively. These actions block each other and both $cts$ and $ctu$ are initially false, so only one of these atoms can be set to true in any plan. That is, $cts$ and $ctu$ are mutually exclusive. Group II consists of actions that all require $cts$ true and group III consists of actions that all require $ctu$ true. These two groups of actions are thus also mutually exclusive. Hence, every plan must start with exactly one action from group I and the rest of the plan consists only of actions from either group II or group III, depending on which the first action is. The intention of this is the following: if the plan starts with action $acs$ then it commits to verifying that $\boldsymbol{s}_n^i$ is satisfiable and if the plan starts with action $acu$ then it commits to verifying that $\boldsymbol{s}_n^i$ is not satisfiable. In either case the plan ends with an action that satisfies the goal only if the plan has verified the commitment made by the first action. This can be interpreted as viewing

| I | $acs\colon \{\overline{ctu}\} \Rightarrow \{cts\}$ | Commit to prove satisfiability |
|---|---|---|
| | $acu\colon \{\overline{cts}\} \Rightarrow \{ctu\}$ | Commit to prove unsatisfiability |
| II | $aset_i\colon \{cts, \overline{v_0}\} \Rightarrow \{x_i\}$ | Set $x_i$ |
| | $avt_0\colon \{cts\} \Rightarrow \{v_0\}$ | Start verification |
| | $avt_j^0\colon \{cts, \overline{e_n^j}, v_{j-1}\} \Rightarrow \{v_j\}$ | Skip disabled clause $c_j$ |
| | $avt_j^k\colon \{cts, e_n^j, v_{j-1}, \ell_j^k\} \Rightarrow \{v_j\}$ | Verify $c_j$ true since $\ell_j^k$ true |
| | $ags\colon \{cts, v_{m(n)}\} \Rightarrow \{goal\}$ | Conclude instance satisfiable |
| III | $avf_j\colon \{ctu, \overline{inc}, e_n^j, \overline{\ell_j^1}, \overline{\ell_j^2}, \overline{\ell_j^3}\} \Rightarrow \{inc\}$ | Verify clause $c_j$ false |
| | $aix_i\colon \{ctu, inc, \overline{x_i}, x_{i-1}, \ldots, x_1\}$ | Increment counter |
| | $\qquad \Rightarrow \{\overline{inc}, x_i, \overline{x_{i-1}}, \ldots, \overline{x_1}\}$ | |
| | $agu\colon \{ctu, inc, x_1, \ldots, x_n\} \Rightarrow \{goal\}$ | Conclude instance unsatisfiable |

Index ranges: $1 \leq i \leq n$, $1 \leq j \leq m(n)$ and $1 \leq k \leq 3$.

Table 1: Actions for Construction 16.

the planner as a theorem prover which first outputs a theorem (the first action in the plan) and then a proof of the theorem (the rest of the plan).

**Lemma 17.** *For every integer $n > 0$ and integer $i$ such that $0 \leq i < 2^{m(n)}$, the* STRIPS *instance $\boldsymbol{p}_n^i$ according to Construction 16 has the following properties:*

1. *It can be computed in polynomial time in $n$.*

2. *It corresponds to the* 3SAT *instance $\boldsymbol{s}_n^i$ such that every plan for $\boldsymbol{p}_n^i$ starts with action $acs$ if $\boldsymbol{s}_n^i$ is satisfiable and otherwise with action $acu$.*

3. *It always has at least one plan.*

*Proof.* Property 1 is trivial to prove. To prove property 2, we first note that the initial state contains only atoms from $E_n$. As previously noted, each subset $E_n^i$ of $E_n$ uniquely identifies the 3SAT instance $\boldsymbol{s}_n^i$ by telling which clauses in $C_n$ are 'enabled' in $\boldsymbol{s}_n^i$.

We have two cases: if the plan starts with action $acs$ then it commits to verifying that $\boldsymbol{s}_n^i$ is satisfiable and if the plan starts with action $acu$ then it commits to verifying that $\boldsymbol{s}_n^i$ is not satisfiable. In either case the plan ends with an action that satisfies the goal only if the plan has verified the commitment made by the first action. The details of the two cases are as follows.

If the plan verifies satisfiability, then it must be of the form

$$\langle acs, \underbrace{aset_{i_1}, \ldots, aset_{i_h}}_{\text{assign}}, avt_0, \underbrace{avt_1^{k_1}, \ldots, avt_{m(n)}^{k_{m(n)}}}_{\text{verify}}, ags \rangle.$$

The assign block has $h$ actions that set a satisfying assignment for $x_1, \ldots, x_n$. The verify block consists of one action $a = avt_j^{k_j}$ for each clause $c_n^j$. If $c_n^j$ is enabled ($e_n^j$ true), then $1 \leq k_j \leq 3$ and $a$ verifies that $\ell_j^k$ in $c_n^j$ is true for the assignment. Otherwise, if $c_n^j$ is disabled then $k_j = 0$, so $a = avt^0_j$ which skips over $c^j_n$ without verifying anything. The planner has thus

1. committed to verify that $\boldsymbol{s}^i_n$ is satisfiable,

2. chosen a satisfying assignment for $x_1, \ldots, x_n$ and

3. chosen one literal for each enabled clause as a witness that the clause is true under this assignment.

The last action $ags$ makes the goal true if these three steps are successful. Note that this works also for the case where no clause is enabled, which corresponds to the trivially satisfiable instance with an empty set of clauses. Obviously, there is a plan of this form if and only if $\boldsymbol{s}^i_n$ is satisfiable.

If the plan instead verifies unsatisfiability, then it must be of the form

$$\langle acu, b_0, a_1, b_1, a_2, b_2, \ldots, a_h, b_h, agu \rangle,$$

where $h = 2^n - 1$. Except for the first and last actions, this plan can be viewed as two interleaved sequences

$$\alpha = \langle a_1, \ldots, a_h \rangle = \langle aix_1, aix_2, aix_1, aix_3, \ldots, aix_1 \rangle$$

and

$$\beta = \langle b_0, b_1, \ldots, b_h \rangle.$$

The $aix_i$ actions are increment actions that use $x_1, \ldots, x_n$ to form a binary counter. Since these variables correspond to the number 0 in the initial state and there are $2^n - 1$ increment actions, the subplan $\alpha$ enumerates all possible truth assignments for $x_1, \ldots, x_n$. Sequence $\beta$ consists only of actions of the type $avf_j$. Each $avf_j$ action verifies that the corresponding clause $c^j_n$ is enabled and false under the current assignment to $x_1, \ldots, x_n$. The $aix_i$ actions require $inc$ to be true and set it false, while the $avf_j$ actions instead require $inc$ to be false and set it true. Hence the plan is synchronised such that it alternates between actions from the two sequences. Since the first $aix_i$ action is preceeded by an $avf_j$ action and there is an $avf_j$ action after the last $aix_i$ action, it follows that there must be some unsatisfied and enabled clause for every possible truth assignment, since the synchronization will otherwise get stuck so the counter cannot increment. That is, there is a plan of this type if and only if $\boldsymbol{s}^i_n$ is unsatisfiable. The last action $agu$ makes the goal true if this succeeded. Note that in this case there is no need for actions to skip over disabled clauses since it is sufficient to demonstrate one enabled clause that is false for each assignment.

It follows that the plan is of the first form if $\boldsymbol{s}^i_n$ is satisfiable and of the second form if $\boldsymbol{s}^i_n$ is unsatisfiable. Furthermore, since the first action is a commitment for the rest of the plan whether to verify satisfiability or unsatisfiability, it is sufficient to check this action to decide if $\boldsymbol{s}^i_n$ is satisfiable or not.

Property 3 follows immediately from property 2 since the plan must be of either of the two forms. □

We now have the necessary tools to prove the theorem.

*Proof of Theorem 15.* Suppose there is an algorithm with either sequential or random access as stated in the precondition of the theorem. We can then solve any 3SAT instance $\boldsymbol{s}_n^i$ in polynomial time by asking the algorithm for the first action of some plan for the corresponding instance $\boldsymbol{p}_n^i$ and tell from this action whether $\boldsymbol{s}_n^i$ is satisfiable. However, this implies that $\mathbf{P} = \mathbf{NP}$. □

This proof would still hold if rewriting Construction 16 as described in Construction 4, that is, Theorem 15 holds even if restricted to the set of unary STRIPS instances only.

## 5. Non-Uniform Compact Representations of Plans

Theorem 15 uses a very strong criterion: it requires that one single algorithm can handle all instances. A more relaxed variant is the non-uniform case, where we allow different representations for different instances. That is, we will consider compact representations of single plans under different access criteria. In order to do so we must first define more precisely what we mean by such representations.

### 5.1 Compact Representations and Access Mechanisms

We define the concepts CSAR and CRAR which are representations of action sequences characterised by their access properties[2].

**Definition 18.** Let $f$ be an arbitrary function. Let $\boldsymbol{f} = \langle V, D, A \rangle$ be an FFP([$\mathbf{P}$]) frame and let $\omega \in A^*$. A *representation* $\rho$ of $\omega$ is a DTM. Furthermore:

1. $\rho$ is *$f$-compact* if $||\rho|| \leq f(||\boldsymbol{f}||)$ and it runs in $f(||\boldsymbol{f}||)$ space including the input and output tapes.

2. $\rho$ is an *$f$-compact sequential-access representation* (*$f$-CSAR*) of $\omega$ if it is $f$-compact, takes no input and generates the actions in $\omega$ sequentially in $f(||\boldsymbol{f}||)$ time for each successive action.

3. $\rho$ is an *$f$-compact random-access representation* (*$f$-CRAR*) of $\omega$ if it is $f$-compact and for an arbitrary index $i$ (where $1 \leq i \leq |\omega|$) as input, it outputs action $i$ of $\omega$ in $f(||\boldsymbol{f}||)$ time.

Note that this definition does not require that the representations are computable. We could have used two separate functions, one to bound the access time and one to bound the size, which would allow for better precision. However, we choose to use a single function for both since this makes the theory simpler and clearer while having sufficient precision for our purposes in this paper. We further consider the output tape as cleared between actions so the output is a single action, not the sequence $\omega$. Also note that the space complexity includes the input and output tapes, which implies that the longest sequence an $f$-CRAR $\rho$ can represent is less than $2^{||\rho||}$ actions since its input is limited to $||\rho||$ bits. A Csar has no corresponding limit since it has no input. Furthermore, the time restriction for an $f$-Csar can be viewed as a generalisation of the polynomial delay concept which is not restricted to polynomials. We will often apply this definition to instances rather than frames. Although this makes a slight difference technically, it is not important in principle and ignoring it allows for simpler theorems and proofs. We write only Crar and Csar when referring to the whole family of representations of a particular type.

---

2. Note that this definition differs slightly from our previous one (Bäckström & Jonsson, 2011b). First, we have generalised the definition to allow compact representations for an arbitrary function $f$, not just an arbitrary polynomial. Second, in order to improve the precision we no longer use the $O(\cdot)$ notation but exact functions. Finally, the representations now have the same restriction for space and time. None of these changes matter for the results in our previous publication, but only for the details of proofs.

## 5.2 Sequential-Access Representations

For sequential access in the non-uniform case we would like to ask if all solvable Strips instances have at least one plan with a polynomial Csar. Unfortunately, that still remains an open question. Hence, we consider a more restricted case of this question where we also require that the Csar must be verifiable within some resource constraint, which we define as follows.

**Definition 19.** For every FFP([**P**]) plan representation type $R$, define the following decision problem:

> Plan Representation Verification
> Instance: An FFP([**P**]) instance $\boldsymbol{p} = \langle V, D, A, I, G \rangle$ and a string $\rho$.
> Question: Is $\rho$ an $R$-representation of a plan for $\boldsymbol{p}$?

The complexity of verification is measured in $||\boldsymbol{p}|| + ||\rho||$. We can now state the following theorem about polynomial Csars.

**Theorem 20.** *Let $C$ be an arbitrary complexity class and $p$ an arbitrary polynomial. If $p$-Csar verification is in $\boldsymbol{NP}^C$ and every solvable Strips instance has at least one plan with a corresponding $p$-Csar, then $\boldsymbol{PSPACE} \subseteq \boldsymbol{NP}^C$.*

*Proof.* Let $p$ be an arbitrary polynomial. Suppose $p$-Csar verification is in $\mathbf{NP}^C$ and every solvable Strips instance has at least one plan with a corresponding $p$-Csar. Let $\boldsymbol{p}$ be an arbitrary Strips instance. We can then decide if $\boldsymbol{p}$ has a plan by guessing a string of length $p(||\boldsymbol{p}||)$ bits and then check if that string is a $p$-Csar for some plan for $\boldsymbol{p}$. This can be done in polynomial time (in $||\boldsymbol{p}||$) using an NTM with an oracle for $C$ since $p$-Csar verification is in $\mathbf{NP}^C$. However, deciding if a Strips instance has a plan is **PSPACE**-complete (Bylander, 1994, Thm. 3.1) so it follows that **PSPACE** $\subseteq \mathbf{NP}^C$, since $\boldsymbol{p}$ was chosen arbitrarily. $\qquad\square$

That is, a Csar for a planning instance is of limited use if we must first verify that it is correct before using it, since this verification may be as difficult as solving the instance itself. Also note that if $C$ is a class in the polynomial hierarchy, then **PSPACE** $\subseteq \mathbf{NP}^C$ implies a collapse of this hierarchy. The preceding theorem holds for the restriction to unary Strips instances, since planning is still **PSPACE**-complete for this restriction (Bylander, 1994, Thm. 3.3). In fact, it holds for all restrictions where planning is still **PSPACE**-complete, which includes several other cases in Bylander's analysis as well as many subclasses of SAS$^+$ planning (see Bäckström & Nebel, 1995; Jonsson & Bäckström, 1998a, for overviews of results).

Although this result may seem disapointing, it holds only under the condition that we must check whether the CSAR is correct. That means, for instance, that the theorem is irrelevant if correctness of the CSAR is guaranteed by design. One such case is the following.

**Theorem 21.** *Every* STRIPS *instance according to Construction 16 has a plan with a polynomial* CSAR.

*Proof.* Consider an arbitrary such instance $\boldsymbol{p}_n^i$. Add $n + 1$ extra bits $b_0, \ldots, b_n$ such that $b_0$ tells if $\boldsymbol{s}_n^i$ is satisfiable or not. If $\boldsymbol{s}_n^i$ is satisfiable then the remaining bits specify a satisfying assignment such that $b_i$ gives the value for $v_i$, and they are otherwise undefined. We claim there is a simple deterministic algorithm that uses only $\boldsymbol{p}_n^i$ and $b_0, \ldots, b_n$ and generates a plan for $\boldsymbol{p}_n^i$ with polynomial delay as follows.

Suppose $b_0$ says that $\boldsymbol{s}_n^i$ is satisfiable and that $h$ of bits $b_1, \ldots, b_n$ are one. Then there is a plan for $\boldsymbol{p}_n^i$ of the form

$$\langle acs, \underbrace{aset_{i_1}, \ldots, aset_{i_h}}_{\text{assign}}, avt_0, \underbrace{avt_1^{k_1}, \ldots, avt_{m(n)}^{k_{m(n)}}}_{\text{verify}}, ags \rangle.$$

The actions in the assign block can easily be generated from $b_1, \ldots, b_n$. For the $avt_j^k$ actions, output $avt_j^0$ if $c_n^j$ is not enabled and otherwise output $avt_j^k$ for the smallest $k$ such that $\ell_j^k$ is true for the specified assignment. Clearly this algorithm works with polynomial delay.

Instead suppose $\boldsymbol{s}_n^i$ is not satisfiable. Then the plan is of the second type and must cycle through all possible assignments. Doing this and generating the corresponding counting actions is trivial. For each assignment we must also output an $avf_j$ action. Determine the smallest $j$ such that $c_n^j$ is enabled but not satisfied for the current assignment, and output $avf_j$. This can be done in polynomial time since there is only a polynomial number of clauses.

Clearly this construction is a polynomial CSAR for some plan for $\boldsymbol{p}_n^i$. $\qquad\square$

The following theorem demonstrates that there are also more general and harder classes of instances where even optimal plans have polynomial CSARs by design.

**Theorem 22.** *There is a subclass $X$ of* STRIPS *and a polynomial $p$ such that deciding if instances of $X$ have a plan is* **PSPACE**-*complete and all solvable instances of $X$ have an optimal plan with a $p$-*CSAR.

*Proof.* **PSPACE** can be characterised by the class of polynomial-space bounded DTMs. Bylander (1994, Thm. 3.1) used this fact to demonstrate a polynomial reduction from **PSPACE** to STRIPS planning. We refer to Bylander for details but in brief: given a machine $M$ with input $x$ he constructs a deterministic STRIPS instance that has a plan if and only if $M(x)$ accepts. Hence, it is a polynomial time problem to check that we are in a valid state and then find the only action, if any, that can be applied in that state. It follows that there is some polynomial $p$ such that every such solvable STRIPS instance has a plan with a $p$-CSAR. Furthermore, since the instance is deterministic it has only one plan, which must then be optimal. $\qquad\square$

An even more general observation is that every deterministic FFP($[\mathbf{P}]$) instance that is solvable has exactly one plan, which is thus optimal, and that this plan has a polynomial CSAR. As a contrast to this we will next consider a class of instances where solvable instances always have plans with polynomial CSARs but where we have no optimality guarantee. This example thus illustrates that a CSAR is just a representation and that it gives no guarantees about the actual data it represents. More precisely, the example uses the class of reversible FFP($[\mathbf{P}]$) instances, where the state-transition graph is symmetric.

**Definition 23.** An FFP($[\mathbf{P}]$) frame $\boldsymbol{f} = \langle V, D, A \rangle$ is *reversible* if for all pairs of states $s$ and $t$ in $S(\boldsymbol{f})$, whenever there is an action $a$ in $A$ from $s$ to $t$ then there is also an action $a'$ in $A$ from $t$ to $s$.

Note that reversible instances are not an easy special case of planning; deciding if there is a plan or not is still **PSPACE**-complete (Jonsson et al., 2000, Thm. 18). That is, plans can still be of exponential length.

**Theorem 24.** *There is a polynomial $q$ such that for all polynomials $p$, every solvable and reversible FFP(p) instance has a $(q \circ p)$-CSAR for some plan.*

*Proof.* Let $\boldsymbol{p} = \langle V, D, A, I, G \rangle$ be a solvable FFP($p$) instance such that $\boldsymbol{f} = \langle V, D, A \rangle$ is reversible. Consider the algorithm in Figure 2. Optplan is assumed to be an algorithm such that optplan($s,G$) returns the length of the shortest plan from $s$ to $G$.

If ignoring process B, it is clear that the algorithm outputs an optimal plan for $\boldsymbol{p}$ since there is a plan by assumption. Process B finds two actions $a_1$ and $a_2$ such that executing $\langle a_1, a_2 \rangle$ in state $s$ ends up in state $s$. Such a choice of actions must exist since there is a plan from $s$ to some goal state and $\boldsymbol{f}$ is reversible. The synchronisation between processes A and B make all such actions $a_1, a_2$ appear adjacent in that order in the output of the algorithm, and they do thus not interfere with the plan produced by process A. They just make the plan longer. Choosing actions in process B can be done by a double loop through all pairs of actions and checking them.

It is obvious that there is a polynomial $r$ such that process B runs in $r(||\boldsymbol{p}||)$ time. We can choose $r$ to also allow extra time to run process A in parallel. Let $\rho$ be $\boldsymbol{p}$ together with the algorithm. Then $||\rho|| \leq ||\boldsymbol{p}|| + c$ for some constant $c$. Choose $r$ such that it also satisfies $n + c \leq r(n)$ for all $n > 0$. Obviously, $\rho$ is an $r$-CSAR for some plan for $\boldsymbol{p}$. Choose the polynomial $q$ such that $r(n) \leq q(p(n))$ for all $n > 0$. □

Algorithm Optplan must obviously run in polynomial space, but its complexity is otherwise not important. The parallel algorithm used in the proof of this theorem is to some extent a nonsense algorithm. Process A does all the job by consulting an ordinary planning algorithm (optplan) but gives no time guarantees. Process B, on the other hand, contributes nothing relevant to the plan but satisfies the access time requirement. That is, process B buys time for process A to find the plan by generating irrelevant actions frequently enough to satisfy its time requirements. The wait statements are not strictly necessary but illustrate that we can tune the step complexity of the algorithm by slowing down process B if desired.

Although this example might, perhaps, be considered somewhat pathological, it clearly demonstrates that a CSAR (just like a CRAR) is only a representation. Just like most other data structures it has certain access properties but does not guarantee any particular

```
1   s := I
2   while G(s) = 0 do
3     do in parallel
4       process A:
5         ℓ = optplan(s, G)
6         for all a' ∈ A s.t. pre(a')(s) = 1 do
7           t := post(a')(s)
8           if optplan(t, G) < ℓ then
9             a := a', ℓ := optplan(t, G)
10      process B:
11        while process A running do
12          choose a₁ ∈ A s.t. pre(a₁)(s) = 1
13          u := post(a₁)(s)
14          choose a₂ ∈ A s.t. pre(a₂)(u) = 1 and post(a₂)(u) = s
15          output a₁
16          wait T
17          output a₂
18          wait T
19    output a
20    s := post(a)(s)
```

Figure 2: CSAR algorithm for reversible instances.

properties of the actual data stored. This not uncommon for plan representations either. For instance, a reactive plan could be constructed to have the same behaviour and still be considered correct and run in polynomial time and space. We can alternatively use a random walk algorithm, which will also output actions with polynomial delay. It will eventually reach the goal if there is a plan, but it will also output a lot of redundant actions. The algorithm in Figure 2 may, in a sense, be viewed as a derandomized variant of random walk.

## 5.3 Random-Access Representations

The case of non-uniform random access is clearer than the case of sequential access. Here we can answer the question of existence for CRARs without any further qualifications about verifiability.

**Theorem 25.** *If there is a polynomial $p$ such that every solvable* STRIPS *instance has at least one plan with a corresponding $p$-*CRAR*, then the polynomial hierarchy collapses.*

Since this theorem is not conditioned by verifiability of the representations it is a stronger result than Theorem 20. Before proving the theorem we need to introduce some additional theory.

**Construction 26.** Let $n > 0$ be an arbitrary integer. Construct a STRIPS instance $\boldsymbol{p}_n = \langle V_n, A_n, \varnothing, \{goal\} \rangle$ such that

$$V_n = X_n \cup E_n \cup \{v_0, \ldots, v_{m(n)}\} \cup \{svi, sva, sia, sii, sti, t, f, goal\}$$

and $A_n$ has the actions specified in Table 2.

| | |
|---|---|
| $abi: \{\overline{svi}, \overline{sva}, \overline{sia}, \overline{sii}, \overline{sti}\} \Rightarrow \{svi, \overline{t}\}$ | Begin instance block |
| $aba: \{svi, \overline{sia}\} \Rightarrow \{sva, \overline{f}, v_0, \overline{v_1}, \ldots, \overline{v_{m(n)}}\}$ | Begin assignment block |
| $avt_j^k: \{sva, \overline{v_j}, v_{j-1}, e_n^j, \ell_j^k\} \Rightarrow \{v_j\}$ | Verify $\ell_j$ true in clause $c_j$ |
| $avf_j: \{sva, \overline{v_j}, v_{j-1}, e_n^j, \overline{\ell_j^1}, \overline{\ell_j^2}, \overline{\ell_j^3}\} \Rightarrow \{v_j, f\}$ | Verify clause $c_j$ false |
| $avs_j: \{sva, \overline{v_j}, v_{j-1}, \overline{e_n^j}\} \Rightarrow \{v_j\}$ | Skip disabled clause $c_j$ |
| $aaf: \{sva, v_{m(n)}, f\} \Rightarrow \{\overline{sva}, sia\}$ | Verify assignment satisfying |
| $aat: \{sva, v_{m(n)}, \overline{f}\} \Rightarrow \{\overline{sva}, sia, t\}$ | Verify assignment not satisfying |
| $aix_i: \{sia, \overline{x_i}, x_{i-1}, \ldots, x_1\} \Rightarrow \{\overline{sia}, x_i, \overline{x_{i-1}}, \ldots, \overline{x_1}\}$ | Increment assignment counter |
| $arx: \{sia, x_n, \ldots, x_1\} \Rightarrow \{\overline{sia}, \overline{svi}, sti, \overline{x_n}, \ldots, \overline{x_1}\}$ | Reset assignment counter |
| $ais: \{sti, t\} \Rightarrow \{\overline{sti}, sii\}$ | Verify instance satisfiable |
| $aiu: \{sti, \overline{t}\} \Rightarrow \{\overline{sti}, sii\}$ | Verify instance unsatisfiable |
| $aii_j: \{sii, \overline{e_n^j}, e_n^{j-1}, \ldots, e_1\} \Rightarrow \{\overline{sii}, e_n^j, \overline{e_n^{j-1}}, \ldots, \overline{e_n^1}\}$ | Increment instance counter |
| $ari: \{sii, e_n^{m(n)}, \ldots, e_n^1\} \Rightarrow \{goal\}$ | All instances checked |

Index ranges: $1 \leq i \leq n$, $1 \leq j \leq m(n)$ and $1 \leq k \leq 3$.

Table 2: Actions for Construction 26.

The previous Construction 16 allows plans of two types, either choosing an assignment and then verifying all clauses by chaining, or enumerating all assignments and demonstrate one false clause for each. Construction 26 mixes these methods. To check if an instance is satisfiable the plan must enumerate all variable assignments and for each assignment it must walk through all clauses by chaining. For each enabled clause it demonstrates either a true literal or that none of the literals is true, while disabled clauses are skipped over. Atoms $f$ and $t$ keep track of whether all clauses were true for some assignment, in which case the instance is satisfiable. An extra counter that uses the variables $e_n^1, \ldots, e_n^{m(n)}$ enumerates all possible subsets $E_n^i$ of $E_n$, thus implicitly enumerating all 3SAT instances $\boldsymbol{s}_n^0, \ldots, \boldsymbol{s}_n^{2^{m(n)}-1}$. This counter constitutes an 'outer loop', so for each $E_n^i$, all possible assignments for $x_1, \ldots, x_n$ are tested as described above. The plan can be thought of as implementing the algorithm in Figure 3.

**Lemma 27.** *For all integers $n > 0$, instance $\boldsymbol{p}_n$ according to Construction 26 has the following properties:*

1. *It can be computed in polynomial time in $n$.*

```
1    for all 3SAT instances s of size n do
2        clear t
3        for all assignments to x_1, ..., x_n do
4            clear f
5            for all clauses c_j do
6                if c_j disabled in s then
7                    do nothing
8                elsif some ℓ_j^k in c_j is satisfied then
9                    do nothing
10               else (neither of ℓ_j^1, ℓ_j^2 or ℓ_j^3 is satisfied)
11                   set f
12           if not f then
13               set t
14       if t then
15           report s as satisfiable
16       else
17           report s as unsatisfiable
```

Figure 3: Algorithmic description of Construction 26.

2. *It always has at least one plan.*

3. *There exist constants $a_n$ and $b_n$ such that for every $i$, where $0 \le i < 2^{m(n)}$, the action at position $b_n i + a_n$ in any plan for $\boldsymbol{p}_n$ is $ais$ if the **3SAT** instance $\boldsymbol{s}_n^i$ is satisfiable and it is $aiu$ if $\boldsymbol{s}_n^i$ is unsatisfiable.*

*Proof.* In addition to the previous explanation of the construction we note that the instance is designed to be deterministic. Setting $a_n = 2^n(m(n) + 3) + 2$ and $b_n = a_n + 1$ satisfies the claim, since the action at position $b_n i + a_n$ is $ais$ if $\boldsymbol{s}_n^i$ is satisfiable and it is $aiu$ if it is unsatisfiable. □

All is now set to prove the theorem.

*Proof of Theorem 25.* Suppose $p$ is a polynomial such that all solvable STRIPS instances have at least one plan with a corresponding $p$-CRAR. For each $n > 0$, let $\boldsymbol{p}_n$ be the corresponding instance according to Construction 26 and let $\rho_n$ be a $p$-CRAR for some plan $\omega_n$ for $\boldsymbol{p}_n$. By assumption, such $\omega_n$ and $\rho_n$ must exist for every $n$.

Construct an advice-taking DTM $M$ which takes input of the form $\boldsymbol{I}_n^i = \langle \boldsymbol{p}_n, i \rangle$, where $n$ and $i$ are integers such that $n > 0$ and $0 \le i < 2^{m(n)}$. Let $i$ be represented in binary using exactly $m(n)$ bits. Then, $||\boldsymbol{I}_n^i||$ is strictly increasing in $n$ and depends only on $n$. Let $s_n = ||\boldsymbol{I}_n^i||$ (which is well defined since $||\boldsymbol{I}_n^i||$ does not depend on $i$). Define the advice function $a$ such that $a(s_n) = \rho_n$. The advice is thus a $p$-CRAR for some plan for the STRIPS instance $\boldsymbol{p}_n$ according to Construction 26. (Recall that we only need to know that the advice exists, not how to find it.) Let $a_n$ and $b_n$ refer to the corresponding constants that must exist according to Lemma 27. Since $M$ can run whatever algorithm is used to access $\rho_n$, it follows from the assumptions that $M$ can find action $b_n i + a_n$ in $\omega_n$ in polynomial time in $s_n$. Let it return yes if this action is $ais$ and otherwise return no.

Given an arbitrary 3SAT instance $\boldsymbol{s}_n^i$, compute the corresponding input $\boldsymbol{I}_n^i$ for $M$ and then run $M$ on this instance. By construction, $M$ answers yes if and only if $\boldsymbol{s}_n^i$ is satisfiable. The input $\boldsymbol{I}_n^i$ can be computed in polynomial time in $||\boldsymbol{s}_n^i||$ and $M$ runs in polynomial time using polynomial advice. Since $M$ solves satisfiability for 3SAT it follows that $\mathbf{NP} \subseteq \mathbf{P/poly}$, which is impossible unless the polynomial hierarchy collapses at level 2 according to Theorem 7a. $\qquad\square$

It is further worth noting that the plans for instances according to Construction 26 contains a subplan for every 3SAT instance of a particular size. Hence, we can alternatively view such a plan as a representation of a set of exponentially many plans, which shows that representing one long plan and representing a large set of plans are not fundamentally different issues.

Also for CRARs there are restricted cases where we can prove that they always exist. One such example is, once again, Construction 16.

**Theorem 28.** *Every* Strips *instance according to Construction 16 has a plan with a polynomial* CRAR.

*Proof.* Add $n+1$ extra bits $b_0, \ldots, b_n$ as explained in the proof of Theorem 21 and construct a polynomial-time random-access algorithm as follows.

Suppose $b_0$ says that $\boldsymbol{s}_n^i$ is satisfiable and that $h$ of bits $b_1, \ldots, b_n$ are one. Then there is a plan for $\boldsymbol{p}_n^i$ of the form

$$\langle acs, \underbrace{aset_{i_1}, \ldots, aset_{i_h}}_{\text{assign}}, avt_0, \underbrace{avt_1^{k_1}, \ldots, avt_{m(n)}^{k_{m(n)}}}_{\text{verify}}, ags \rangle.$$

Since $h \leq n$ we can construct the whole assign sequence and then determine a specific action in it in polynomial time. Each of the $avt_j^k$ actions correspond to a specific clause $c_n^j$. Since the clauses are ordered we can compute $j$ from the index of the action we ask for. If $c_n^j$ is not enabled then output $avf_j^0$ and otherwise output $avf_j^k$ for the smallest $k$ such that $\ell_j^k$ is true for the specified assignment $b_1, \ldots, b_n$.

Instead suppose $\boldsymbol{s}_n^i$ is not satisfiable. Then all but the first and last actions are interleaved $aix_i$ and $avf_j$ actions. The $aix_i$ actions have the function of a counter over the variables $x_1, \ldots, x_n$. Let $i$ be an arbitrary index $i$ into the plan where $i$ is not the first or last action. If $i$ is odd, then $a_i$ is an $aix_k$ action and $k$ can easily be computed from $i$, so output $aix_k$. If $i$ is even then $a_i$ is an $avf_j$ action. The value of $x_1, \ldots, x_n$ immediately before $a_i$ can easily be computed from $i$. Use this value and check all enabled clauses in order until finding a clause $c_n^j$ that is not satisfied and then output $avf_j$.

Clearly this construction is a polynomial CRAR for some plan for $\boldsymbol{p}_n^i$. $\qquad\square$

Another and much larger class of instances where plans have CRARs will be considered in the next section.

## 5.4 Relationships between Compact Representations

In this section we investigate how the CRAR and CSAR concepts relate to each other. Since a macro plan can also be viewed as a compact representation we also investigate how macro plans relate to our concepts. We start by showing that if there is a polynomial CRAR for a plan then there is also a polynomial CSAR for that plan.

**Theorem 29.** *There is a polynomial $q$ such that for all polynomials $p$, all FFP([**P**]) frames $\boldsymbol{f} = \langle V, D, A \rangle$ and all $\omega \in A^*$, if $\omega$ has a $p$-CRAR, then it also has a $(q \circ p)$-CSAR.*

*Proof.* Let $\boldsymbol{f}$ be arbitrary FFP([**P**]) instance and let $\rho$ be a $p$-CRAR for some plan $\omega$ for $\boldsymbol{f}$. Use an action counter that is initiated to index the first action in the plan. Generate the actions of the plan sequentially by repeatedly asking $\rho$ for the action indexed by the counter and then incrementing the counter. Let $\rho'$ denote this algorithm together with $\rho$. A $||\rho||$-bit counter is sufficient since $\omega$ must be shorter than $2^{||\rho||}$ actions. Suppose it takes $r(m)$ time to increment an $m$-bit counter. Then there is a constant $c$ such that $||\rho'|| \leq ||\rho|| + c$, $\rho'$ runs in $2p(||\boldsymbol{f}||) + c$ space and $\rho'$ runs in $p(||\boldsymbol{f}||) + r(p(||\boldsymbol{f}||)) + c$ time. Define the polynomial $q$ as $q(n) = 2r(n) + c$. Obviously, $||\rho'|| \leq ||\rho|| + c \leq q(||\rho||) \leq q(p(||\boldsymbol{f}||))$, $2p(||\boldsymbol{f}||) + c \leq q(p(||\boldsymbol{f}||))$ and $p(||\boldsymbol{f}||) + r(p(||\boldsymbol{f}||)) + c \leq q(p(||\boldsymbol{f}||))$ so $\rho'$ is a $(q \circ p)$-CSAR for $\omega$. □

The opposite of this does not hold, however. In particular, while not all instances according to Construction 26 have a plan with a CRAR they all have a plan with a CSAR, so this construction acts as a separation between the two concepts.

**Theorem 30.** *Unless the polynomial hierarchy collapses, there is no polynomial $q$ such that for every polynomial $p$, every STRIPS instance $\boldsymbol{p}$ and every plan $\omega$ for $\boldsymbol{p}$, if $\omega$ has a $p$-CSAR then $\omega$ has a $(q \circ p)$-CRAR.*

*Proof.* Let $X$ denote the class of STRIPS instances used in the proof of Theorem 25. Since these instances are deterministic there is a polynomial $p$ such that every solvable instance has a $p$-CSAR for some plan. However, it follows from the same proof that there is no polynomial $r$ such that all instances of $X$ have a plan with an $r$-CRAR. Hence, there is no polynomial $q$ such that all instances have a plan with a $(q \circ p)$-CRAR. □

Although not previously defined in the paper, it makes sense to also have a look at macro plans in this context. A macro is a sequence of two or more actions. Macros are commonly used in planning and treated as a single action by the planner. Macros can be useful for planning if there are certain sequences of actions that occur frequently in plans (Korf, 1987). However, macros may also be used for the purpose of representing a plan in a more compact and structured way. This is especially true if macros are allowed to also contain other macros, since this allows hierarchies of macros. For instance, it is well known that the shortest solution for the Towers-of-Hanoi problem for arbitrary number of disks can be described by a recursive schema (Gill, 1976, Ex. 3–19) although the plan itself is exponential in the number of disks. The 3S class of planning instances (Jonsson & Bäckström, 1998b) has the property that we can always find out in polynomial time if an instance has a plan, but the plan itself may be of exponential length and thus cannot be generated in subexponential time. Giménez and Jonsson (2008) showed that plans for 3S instances always have a polynomial-size representation using macros. In fact, such a macro plan can even be generated in polynomial time although the actual non-macro plan would take exponential time to generate. This result was later generalised to some other classes of planning instances by Jonsson (2009). We will show that polynomial-size macro plans have an immediate connection to compact plan representations. However, in contrast to Giménez and Jonsson we will not discuss how to generate macro plans but only analyse some of their properties.

Macro plans can be very powerful tools for representing plans compactly. Hence, it is interesting to identify criteria for when compact macro-plan representations exist and not. That problem is out of the scope of this paper, but we will give a partial answer to the question in the following way. It is straightforward to see that a macro plan can be viewed as a context free grammar (CFG): let the actions be the terminals, let the macros be the variables, let the macro expansions be the production rules and let the root macro be the start symbol. We note that if we use macros to represent a single plan, rather than to represent various possibilities for planning, then the macro expansions must be acyclic in order to produce a unique well-defined plan. Hence, a macro plan can be defined as an acyclic CFG. When such CFGs are used to represent a single string compactly they are often referred to as compressed grammars. Furthermore, such a compressed grammar permits efficient random access into the string it represents; both the access and the necessary preprocessing is polynomial time in the size of the grammar (Bille et al., 2011). More precisely, consider a grammar of size $n$ that represents a string of length $N$ with a derivation tree of maximum height $h$. After a polynomial time preprocessing, in the size of the grammar, it is possible to random access any symbol in the string by index in $O(\log N)$ time or, alternatively, in $O(h)$ time. Such algorithms typically work by first computing the length of the substrings generated by each rule, the preprocessing step, and then use this information to find the symbol with a certain index by top-down search. Since the grammar is acyclic we get $h \leq n$. Hence, the following proposition is immediate from the properties of compressed grammars.

**Proposition 31.** *There is a polynomial $r$ such that for every FFP([**P**]) frame $\langle V, D, A \rangle$ and every macro plan $\mu$ for a sequence $\omega \in A^*$, $\mu$ can be used to random access any action in $\omega$ in $r(||\mu||)$ time.*

We thus get the following relationship between macro plans and CRARs.

**Theorem 32.** *There is a polynomial $p$ such that for all polynomials $q$, all FFP([**P**]) frames $\boldsymbol{f} = \langle V, D, A \rangle$ and all action sequences $\omega \in A^*$, if $\omega$ has a macro plan $\mu$ such that $||\mu|| \leq q(||\boldsymbol{f}||)$ then $\omega$ has a $(p \circ q)$-CRAR.*

*Proof.* Let $r$ be a polynomial such that all macro plans $\mu'$ can be random accessed in $r(||\mu'||)$ time. Let $\rho$ be $\mu$ together with the random access algorithm. Then $||\rho|| \leq ||\mu|| + c$ for some constant $c$. Define $p$ such that $p(n) = r(n) + c$. We get $||\rho|| \leq ||\mu|| + c \leq q(||\boldsymbol{f}||) + c \leq r(q(||\boldsymbol{f}||)) + c = p(q(||\boldsymbol{f}||))$. Furthermore, $\rho$ runs in $r(||\mu||) \leq r(q(||\boldsymbol{f}||)) + c = p(q(||\boldsymbol{f}||))$ space and time. It follows that $\rho$ is a $(p \circ q)$-CRAR for $\omega$. □

It follows from Theorem 29 that every plan with a polynomial macro plan also has a polynomial CRAR. That is, the class of polynomial macro plans is a subclass of the class of polynomial CRARs, but we do not know if it is a proper subclass. In any way, these results do imply that we cannot always find a polynomial macro plan for an instance.

**Corollary 33.** *If there is a polynomial $p$ such that every solvable STRIPS instance has at least one plan with a corresponding macro plan of size $p(||\boldsymbol{p}||)$ then the polynomial hierarchy collapses.*

*Proof.* Immediate from Theorems 25 and 32. □

## 6. Problem Reformulation

Having now concluded that there seems to be little hope that plans can be compactly represented in the general case, we turn to the idea of problem reformulation to see if that can be of any help. While this may seem out of place in this context it is, to the contrary, a quite logical step to take. So far, we have only analysed planning problems and plans, and that is what the results hold for. It is not obvious that, or when, the results hold also when planning instances are solved by reformulating them to instances of some other problem. It is thus hypothetically possible that we could get around the problems with this approach. However, to say something useful and relevant about this, it is not sufficient to look only at naive approaches, such as polynomial reductions, so we will investigate a stronger criterion.

The basic idea of reformulation is to transform a planning instance to another equivalent instance, either another planning instance or an instance of some other problem. For reformulation to be useful, the solution for the new instance must be of use to solve the original instance, and something must be gained. Often, reformulation is used with the intention that the overall process is faster than solving the original instance directly. Common variants are to reformulate planning into SAT, CSP, model checking or another planning problem. Reformulation of planning into SAT was first suggested by Kautz and Selman (1992) and is still a popular approach to planning. Long, Fox, and Hamdi (2002) discuss reformulation for planning in general and Edelkamp, Leue, and Visser (2007) discuss the connections between model checking and planning.

The reformulation process can be viewed as shown in Figure 4. A planning instance $\boldsymbol{p}$ has a solution $\omega$ that we can find directly using ordinary planning. Solving $\boldsymbol{p}$ via reformulation instead follows the indirect path in the figure. First $\boldsymbol{p}$ is reformulated into a new instance $R(\boldsymbol{p})$ (of some problem). Then this instance is solved which produces a solution $\rho$ for $R(\boldsymbol{p})$. Finally, $\rho$ is transformed back into a solution $\omega$ for $\boldsymbol{p}$.

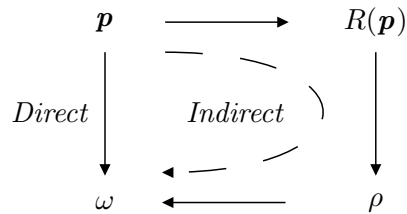

Figure 4: Reformulation of the generation problem.

Obviously, reformulation cannot help us when plans are exponential. Even if the first two steps of the indirect path took polynomial time and $\rho$ was of polynomial size, it would still necessarily take exponential time to transform $\rho$ into $\omega$ because $\omega$ is exponential. That is, the problem is inherently intractable whichever method we use to solve it. Reformulation could potentially speed things up, if $\rho$ could somehow be used directly as a solution for the original problem, but that would happen rarely, if at all.

The situation is different, though, if we consider the decision problem rather than the generation problem, that is, if we ask not for a plan but for whether there is a plan or not. In this case we can use the solution for $R(\boldsymbol{p})$ directly, since decision problems have only two possible answers, yes or no. We may thus escape the inherent intractability. This variant of reformulation is shown in Figure 5. Since no exponential solution is generated in this case, reformulation could potentially be more efficient. We know that the decision problem for STRIPS is **PSPACE**-complete in the general case. If the reformulated problem were easier to solve, then it could be beneficial to first reformulate $\boldsymbol{p}$ to $R(\boldsymbol{p})$ and ask if that instance has a solution or not. Then it would be possible to check if there is a solution at all before embarking on generating a possibly exponentially long plan. Consider, for instance, the 3S class (Jonsson & Bäckström, 1998b) where plans may be of exponential size but it is always possible to decide in polynomial time if there is a plan. It thus seems like the case of reformulating decision problems is the most interesting one to look at, and if that does not give any improvement, then there can hardly be any improvement for plan generation via reformulation either.

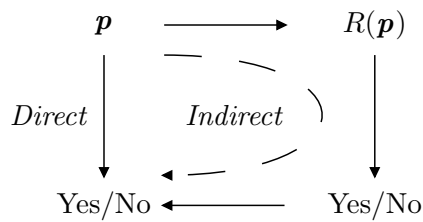

Figure 5: Reformulation of the decision problem.

Let $\mathsf{PE}(\textsc{Strips})$ denote the decision problem (that is, plan existence) for STRIPS. The following two results are trivial, but illustrative.

**Theorem 34.** *a) There exists a decision problem $\mathsf{X}$ and a function $R$ such that it holds for all $\boldsymbol{p} \in \mathsf{PE}(\textsc{Strips})$ that $R(\boldsymbol{p}) \in \mathsf{X}$ and that $\boldsymbol{p}$ and $R(\boldsymbol{p})$ have the same answer. b) If there is some complexity class $C$, some decision problem $\mathsf{X} \in C$ and a polynomial-time computable function $R$ such that it holds for all $\boldsymbol{p} \in \mathsf{PE}(\textsc{Strips})$ that $R(\boldsymbol{p}) \in \mathsf{X}$ and that $\boldsymbol{p}$ and $R(\boldsymbol{p})$ have the same answer, then $\mathbf{PSPACE} \subseteq C$.*

*Proof.* a) Let $\mathsf{X} = \mathsf{PE}(\textsc{Strips})$ and $R$ the identity function. b) Immediate, since $R$ is a polynomial reduction from $\mathsf{PE}(\textsc{Strips})$ to $\mathsf{X}$. $\qquad\square$

In both cases we reformulate a **PSPACE**-complete problem into a **PSPACE**-complete problem, which is not very interesting. If we are to prove anything better, we must obviously look for an $\mathsf{X}$ and an $R$ with more useful restrictions.

It is important to note that when reformulating planning into some **NP**-complete problem, for instance SAT, this does not magically make planning **NP**-complete. The reason that STRIPS planning is **PSPACE**-complete is that it allows exponential solutions. As soon as we restrict the solutions to be bounded by some fixed polynomial, planning belongs in **NP**. Furthermore, encodings of planning instances in SAT typically use atoms to encode what actions appear at each position in the plan, that is, an exponential number of extra atoms are required in the general case. Hence, either our original problem was already in **NP** or we have to blow up the instance exponentially when reformulating to SAT. In the latter case, the complexity results are no longer comparable. Also note, that if we deliberately restrict ourselves to ask only if there is a plan of a certain length or shorter, then we are actually solving a restricted version of the optimization problem, and also in this case planning itself would be no harder. In fact, it seems most unlikely that planning in general could be reformulated into a problem in **NP**. In order to avoid straightforward and naive approaches to reformulation we consider and analyse reformulations defined as follows.

**Definition 35.** Let $\boldsymbol{p} = \langle V, A, I, G \rangle$ be a STRIPS instance, let $\boldsymbol{f} = \langle V, A \rangle$ and let $\boldsymbol{d} = \langle I, G \rangle$. Let X be some decision problem. A *reformulation* of PE(STRIPS) into X is a pair $\langle R, r \rangle$ of functions that maps every instance $\boldsymbol{p} = \langle \boldsymbol{f}, \boldsymbol{d} \rangle \in$ PE(STRIPS) to a corresponding instance $\boldsymbol{x} = R(r(\boldsymbol{f}), \boldsymbol{d}) \in$ X such that $\boldsymbol{p}$ and $\boldsymbol{x}$ have the same answer. $\langle R, r \rangle$ is a *polynomial reformulation* if there are also some fixed polynomials $p, q$ such that

    1) $||r(\boldsymbol{f})|| \in O(p(||\boldsymbol{f}||))$ and

    2) $R$ is computable in $O(q(||r(\boldsymbol{f})|| + ||\boldsymbol{d}||))$ time.

We thus consider a reformulation that involves two functions, $R$ and $r$. Function $r$ is the main reformulation function, intended to reformulate the difficult part of the instance. We do not even require this function to be computable, we only require that it exists. Function $R$ is then used to transform the initial and goal descriptions into something similar that the new instance can use, and combine this with the result delivered by $r$ into a proper instance of X. It should be noted that this reformulation concept is similar, although not identical, to the compilation concept used by Nebel (2000).

**Theorem 36.** *There is no polynomial reformulation of PE(*STRIPS*) to some* X $\in$ **NP**, *unless the polynomial hierarchy collapses.*

*Proof.* Suppose $\langle R, r \rangle$ is such a reformulation. For arbitrary integer $n > 0$, let $\boldsymbol{f}_n^u = \langle V_n, A_n \rangle$ be defined as in Construction 16, but without action $acs$, and let $\boldsymbol{d}_n^i = \langle E_n^i, \{goal\} \rangle$, for all $i$ such that $0 \leq i < 2^{m(n)}$. It follows trivially from the proof of Lemma 17 that instance $\boldsymbol{p}_n^i = \langle \boldsymbol{f}_n^u, \boldsymbol{d}_n^i \rangle$ has a solution if and only if $\boldsymbol{s}_n^i$ is unsatisfiable (note that the 'SAT part' of the instance is 'disarmed').

Construct an advice-taking NTM $M$ with input $\boldsymbol{I}_n^i = \langle \boldsymbol{f}_n^u, i \rangle$, for all $n > 0$ and $0 \leq i < 2^{m(n)}$, representing $i$ in binary using $m(n)$ bits. Clearly, $||\boldsymbol{I}_n^i||$ is strictly increasing and depends only on $n$, so let $s_n = ||\boldsymbol{I}_n^i||$ (for arbitrary $i$). Define the advice function $a$ such that $a(s_n) = r(\boldsymbol{f}_n^u)$. (Note that we only need to know that the advice exists, not how to find it). Let $M$ first compute $\boldsymbol{d}_n^i$ from $\boldsymbol{I}_n^i$, and then compute $\boldsymbol{x}_n^i = R(a(s_n), \boldsymbol{d}_n^i) = R(r(\boldsymbol{f}_n^u), \boldsymbol{d}_n^i)$, both in polynomial time since $a(s_n)$ is given for free as advice. By assumption, $\boldsymbol{x}_n^i \in$ X and has answer yes if and only if $\boldsymbol{p}_n^i$ has a solution. Also by assumption, we have X $\in$ **NP** so $M$ can solve $\boldsymbol{x}_n^i$ by guessing a solution and verifying it in polynomial time. Hence, deciding if $\boldsymbol{p}_n^i$ has a solution is in $\mathbf{NP/poly}$.

For an arbitrary 3SAT instance $\boldsymbol{s}_n^i$, compute $\boldsymbol{I}_n^i$ in polynomial time. $M$ answers yes for $\boldsymbol{I}_n^i$ if and only if $\boldsymbol{s}_n^i$ is unsatisfiable. However, unsatisfiability for 3SAT is co$\mathbf{NP}$-complete so it follows that co$\mathbf{NP} \subseteq \mathbf{NP/poly}$, which is impossible unless the polynomial hierarchy collapses to level 3, according to Theorem 7b. $\qquad\square$

This result can be pushed arbitrarily high up in the polynomial hierarchy, thus making it unlikely that planning could be reformulated to anything simpler at all.

**Corollary 37.** *There is no polynomial reformulation $\langle R, r \rangle$ of $PE(\text{STRIPS})$ to some decision problem $\mathsf{X} \in \boldsymbol{\Sigma}_k^p$, for $k > 1$, unless the polynomial hierarchy collapses to level $k + 2$.*

*Proof sketch.* Construction 16 demonstrates how to encode both existential quantification (choosing a truth assignment in the sat part) and universal quantification (enumerating all truth assignments in the unsat part). Hence, it is straightforward to modify it to an analogous construction for QBF formulae with $k$ alternations. Given that, the rest of the proof is analogous to the proof of Theorem 36, but $M$ must use an oracle for $\boldsymbol{\Sigma}_{k-1}^p$. The same argument leads to $\boldsymbol{\Pi}_k^p \subseteq \boldsymbol{\Sigma}_k^p/\mathbf{poly}$, which is impossible unless the polynomial hierarchy collapses to level $k + 2$, according to Theorem 7b. $\qquad\square$

Since both these proofs build on Construction 16 and do not rely the exact position of actions it follows that also this Theorem and Corollary hold when restricted to unary STRIPS instances only.

## 7. Discussion

This section consists of five parts. We first transfer the reformulation theorem to a more general result about adding information to guide planners, and discuss how that can explain various results in the literature. We then discuss the potential relationship between causal graphs and compact representations. This is followed by a discussion on how the results in the paper could be relevant for plan explanation. The fourth part discusses some related work on compact representations and compilation. The section ends with a summary of the results and a list of open questions.

### 7.1 Reformulation and Additional Information

Theorem 36 has broader consequences than just for reformulation. In fact, it implies that there is no way to help a planner by adding information to a planning frame, no matter what information or how we get it, unless we accept that the amount of information is not always polynomially bounded in the frame size. In the following theorem the function $g$ is assumed to represent the additional information, and it need not even be computable. We only require that its result is polynomially bounded.

**Theorem 38.** *Let $p$ be an arbitrary polynomial. Consider a function $g$ and an algorithm $\mathcal{A}$ such that*

    *1. $g$ maps STRIPS frames to $\{0, 1\}^*$ such that $||g(\boldsymbol{f})|| \in O(p(||\boldsymbol{f}||))$ for all frames $\boldsymbol{f}$ and*

2. *for all* STRIPS *instances* $\boldsymbol{p} = \langle \boldsymbol{f}, \boldsymbol{d} \rangle$, *algorithm* $\mathcal{A}$ *answers yes for input* $\langle \boldsymbol{p}, g(\boldsymbol{f}) \rangle$ *if and only if* $\boldsymbol{p}$ *has a plan.*

*If* $\mathcal{A}$ *runs in polynomial time, then the polynomial hierarchy collapses.*

*Proof.* Assume there is a function $g$ and an algorithm $\mathcal{A}$ with the properties described in the theorem. Define a function $r$ such that $r(\boldsymbol{f}) = \langle \boldsymbol{f}, g(\boldsymbol{f}) \rangle$ for every STRIPS frame $\boldsymbol{f}$. Also define a function $R$ such that $R(\langle \boldsymbol{f}, x \rangle, \boldsymbol{d}) = \langle \langle \boldsymbol{f}, \boldsymbol{d} \rangle, x \rangle$ for every STRIPS instance $\boldsymbol{p} = \langle \boldsymbol{f}, \boldsymbol{d} \rangle$ and every string $x$. Then $R(r(\boldsymbol{f}), \boldsymbol{d}) = \langle \langle \boldsymbol{f}, \boldsymbol{d} \rangle, g(\boldsymbol{f}) \rangle = \langle \boldsymbol{p}, g(\boldsymbol{f}) \rangle$, so $\langle R, r \rangle$ is a polynomial reformulation of STRIPS planning into an equivalent problem that algorithm $\mathcal{A}$ can solve in polynomial time. However, no such reformulation can exist according to Theorem 36, unless the polynomial hierarchy collapses. $\square$

This result can be extended upwards in the polynomial hierarchy in the same way as Corollary 37 (no longer requiring $\mathcal{A}$ to be a polynomial algorithm). That means that we cannot make planning simpler by adding a polynomial amount of additional information to a frame and use a clever algorithm to use that information when planning. Planning will remain as hard as it is without that extra information. While it may sometimes help to add information to a particular instance to somehow guide the planner, there is no systematic way to add such information on the frame level if it is required to be of polynomial size.

The planning literature is rich with methods that are intended to make planning more efficient by adding information in one way or another, although the methods are perhaps not always thought of as doing so. A non-exhaustive list of such methods, and similar, is abstraction hierarchies, macros, case-based planning, annotated planning and landmarks.

State space abstraction in planning goes back at least to the ABSTRIPS planner (Sacerdoti, 1974). The main idea is to form abstraction hierarchies on the variables, and thus implicitly on the actions, such that the planner can plan for the most important goals first to get an abstract plan that can then be refined into a more detailed plan. Knoblock (1994) proposed an algorithm for automatically computing such abstraction hierarchies. While his algorithm was successful on many examples it was demonstrated to sometimes fail and produce exponential plans for instances that have a linear optimal plan (Bäckström & Jonsson, 1995). This is not surprising since the use of an abstraction hierarchy can be viewed as adding information to the planning frame. Automatic generation of abstraction hierarchies is a systematic way to add information and can thus be treated as a special case of Theorem 38.

Adding a set of macros to a planning frame is very similar to using abstraction hierarchies, as Knoblock (1993, pp. 110–111) noted. A planner that uses abstraction searches for a plan in an abstract space and then tries to refine each action into a subplan on the lower lever. A planner that uses macros does not search an abstract space but instead already has a set of macros available that each correspond to a subplan. Finding a macro that works and can be expanded is thus very similar to refining an abstract action. Also the use of macros has been demonstrated to speed up planning considerably in certain cases (Korf, 1987). Macros are typically added on the frame level, and learning has been suggested as one method to create macros automatically (Korf, 1985). However, macros are typically treated as any other action by the planner and are not expanded until after finding a plan. Hence, the addition of macros may also backfire and make planning less efficient, just as adding redundant actions may do (Haslum & Jonsson, 2000). Once again, this is not surprising since the addition of macros is addition of information and is thus also covered by Theorem 38.

Case based planning (see Spalazzi, 2001, for a survey) uses stored plans or plan skeletons that the planner tries to reuse by modifying and/or extending them. In one sense, this is similar to macro planning, but with more advanced macros and macro expansion methods. One can also view it as similar to abstraction, where a plan must be refined in order to work. The difference is that the abstraction planner finds the plan skeleton by planning in an abstract space while the case-based planner has a set of such plans stored in a database. These plans may be handcoded, but are usually the result of learning from previous planning situations. It is well known that also case-based planning may fail to improve efficiency and that the cases used must be similar to the actual instance at hand (Nebel & Koehler, 1995; Liberatore, 2005b). Also this can be explained as a special case of our Theorem 38.

The term annotated planning is sometimes used to refer to a number of similar techniques of adding control information to a planner. Examples are the PRODIGY planner (Veloso et al., 1995) which allows control information like rules for goal ordering and TLPLAN (Bacchus & Kabanza, 2000) which allows adding temporal-logic axioms to control the planner. While such techniques can be good if using hand-tailored control rules/axioms for a particular application domain, it is immediate from Theorem 38 that they cannot help us in the general case.

Planning with landmarks (Hoffmann, Porteous, & Sebastia, 2004) is the idea of adding explicit subgoals (called landmarks) to a planning instance. The intention is to tell the planner that the landmarks must be achieved by the plan in order to achieve its overall goal. Landmarks may also be ordered, to further guide the planner. However, as the authors themselves point out, deciding if a variable value (or logic formula) is a necessary subgoal is itself a **PSPACE**-complete problem. Hence, one usually considers incomplete sets of landmarks. More interestingly, landmarks differ from the previous methods above in a very important aspect; landmarks are added on the instance level, not on the frame level. Although this might not be quite a rigid difference in practice, it seems to be fundamental in essence. Hence, adding landmarks is a non-uniform case of adding information and it is thus not immediately covered by Theorem 38. How to meaningfully analyse the non-uniform case remains an open question.

## 7.2 Causal Graphs

Knoblock (1994) defined an ordering on the variables of a planning instance which he used as a guidance for finding abstraction hierarchies. An ordering on the variables was fundamental also for the 3S class (Jonsson & Bäckström, 1998b) and this ordering implicitly defined an abstraction hierarchy. The concept of an ordering on the variables with the intention of defining an abstraction hierarchy, define tractable subclasses etc. is nowadays usually referred to as a *causal graph* (see Chen & Giménez, 2010, for a survey of using properties of the causal graph to define tractable subclasses of planning). Many papers still use Knoblock's definition, which is as follows:

For every STRIPS action $a$ let $V_{\mathrm{pre}(a)} = Atoms(\mathrm{pre}(a))$ and $V_{\mathrm{post}(a)} = Atoms(\mathrm{post}(a))$. Let

$\boldsymbol{f} = \langle V, A \rangle$ be a STRIPS frame. The causal graph for $\boldsymbol{f}$ is the directed graph $G_{CG} = \langle V, \leq \rangle$ where for all $u, v \in V$, $u \leq v$ if and only if both $u \neq v$ and there is some $a \in A$ such that $u \in V_{\text{pre}(a)} \cup V_{\text{post}(a)}$ and $v \in V_{\text{post}(a)}$.

The idea behind causal graphs is that each strongly connected component of the graph should correspond to an abstraction level. If applying this definition to examples in this paper, we find that an instance according to Construction 16 has a causal graph containing a large strongly connected component. That is, it would not be possible to form any good abstraction hierarchies for it based on such causal graphs. However, Theorem 25 says that plans for such instances seem not very likely to have useful compact representations anyway. Plans for the binary counter in Construction 5 do have polynomial CRARs since they have polynomial macro plans. Yet, the whole causal graph for such an instance is also strongly connected. On the other hand, plans for the Gray counter in Construction 6 are exponential and have polynomial CRARs too, but the causal graph is acyclic in this case. It thus seems that the causal graph of the type used by Knoblock and many others is not a sufficient, or even necessarily useful, tool for judging when plans have compact representations. There are other variants of causal graphs, though. One example is interaction networks (Chen & Giménez, 2010). Another is Jonsson's (2009) refined version of Knoblock's causal graph, defined as follows:

Let $\boldsymbol{f} = \langle V, A \rangle$ be a STRIPS frame. The refined causal graph for $\boldsymbol{f}$ is the directed graph $G_{RCG} = \langle V, \leq \rangle$ where for all $u, v \in V$, $u \leq v$ if and only if $u \neq v$ and either

    1) there is some $a \in A$ such that $u \in V_{\text{pre}(a)} - V_{\text{post}(a)}$ and $v \in V_{\text{post}(a)}$ or

    2) there is some $a \in A$ such that $u, v \in V_{\text{post}(a)}$ and either

        a) there is some $a' \in A$ such that $u \in V_{\text{post}(a')}$ and $v \notin V_{\text{post}(a')}$ or

        b) there is no $a' \in A$ such that $u \notin V_{\text{post}(a')}$ and $v \in V_{\text{post}(a')}$.

The major difference between this variant and Knoblock's is that if two variables both appear in the postcondition of the same action, then they do not necessarily form a cycle in the graph. Hence, unary actions is no longer a prerequisite for acyclic graphs. If using the refined causal graph, then both the Gray counter and the binary counter have acyclic graphs, while Construction 16 still has a large strongly connected component. That is, in these three examples acyclicity of the refined causal graph correlates with whether plans have compact representations or not. While this correlation seems not to hold in general, the difference between the two types of causal graphs suggests that further study of variations on the concept could lead to further insight into the topic of compact representations. Should this turn out to be fruitful, then it would likely carry over also to other areas where causal graphs have been used, like model checking (Wehrle & Helmert, 2009).

## 7.3 Plan Explanation

The results in this paper are also important for plan explanation. Bidot et al. (2010) suggest that it is important for planning systems (and other AI systems) to be able to explain their plans and decisions to the user, or else the user may not trust the system. Similarly, Southwick (1991) writes:

> There seems to be a general agreement amongst those involved in KBS research that in order to be useful, a system must be able to explain its reasoning to a user.

Although we do not consider any advanced explanation methods, as they do, our results have implications for what is possible to explain meaningfully. For plan explanation, our results are not necessarily as bad as for planning. Consider for example a plan for an instance of Construction 16. In the case where the 3SAT instance is unsatisfiable, almost the whole plan consists of an alternating sequence of the form $\langle a, b, a, b, a, b, \ldots \rangle$, where $a$ denotes either of the actions $aix_1, \ldots, aix_n$ and $b$ denotes either of the actions $avf_1, \ldots, avf_m$. The first group are actions that together implement an increment function, and thus all serve the same purpose. Similarly, the second group consists of actions that all serve the purpose of verifying that some clause is false. An abstraction of this action sequence could have the form $\langle inc, vfy, inc, vfy, inc, vfy, \ldots \rangle$, where $inc$ denotes any of the counting actions and $vfy$ any of the verification actions. For the purpose of explanation, it seems useful to replace the actual actions with such abstract explanations of their functions. This abstract sequence is easier to understand, and it also allows using macros to compress it, which might further enhance its explaining power. However, in this particular case, it would probably be even more useful to abstract the whole sequence into a for loop, or similar. This essentially boils down to partitioning the set of actions into equivalence classes such that each such class consist of actions that can be meaningfully seen as implementing the same concept. It seems both interesting and important to investigate how and when one can partition the set of actions into equivalence classes useful for such abstractions.

Plan explanation could also mean trace explanation in model checking, where we would analogously make a long trace shorter and more abstract in order to make it easier to understand. It is well known that there are close ties between planning and model checking, and that model-checking traces can be viewed as plans and vice versa (Edelkamp et al., 2007). The number of steps (or clock cycles) can be exponential in the number of state variables; even if the system is divided into subsystems, individual subsystems may have exponential behaviour which blows up when combined with other subsystems. An exponential-size plan/trace is not of much use to an engineer—it is an almost impossible task to analyse and understand such a plan. If the planning/verifying system could autonomously find repetitive patterns, and even recursive repetitive patterns, in the plan and abstract these, then it would be considerably easier to understand what happens and why. In fact, it may not be interesting to execute the plan, even in a simulator, so a compact understandable explanation of the plan may be the actual goal.

Furthermore, Geib (2004) discusses the problem of combinatorial explosion in plan recognition, where an exponential number of plans may share the plan prefix recognized so far. It could clearly be useful to have structured compact representations of plan candidates both to save space and to allow for more intelligent operations on these plans. Although this problem is slightly different from representing a single long plan, we have seen that these two problems are related.

In all these cases, the primary purpose of a compact representation would thus be to find and exploit some inherent structure in the plan, or set of plans, rather than to save space.

## 7.4 Additional Related Work

Liberatore (2005a) has also studied the problem of representing plans compactly and there are similarities as well as differences between his results and ours. In contrast to us, he considers also plans represented as sequences of states, not only sequences of actions. For both cases, he considers a random access representation as well as a sequential representation. His random-access representation of action sequences (TA) is essentially the same as our CRAR concept, except that he specifies that it must be implemented by a circuit. The sequential representation of action sequences (SA), on the other hand, is different from our CSAR concept. It is a function that takes a state as input and returns the next state. Hence, it is more like a restricted type of reactive plan than a CSAR, and his results are thus not immediately comparable to ours. For instance, contrary to our Theorem 29 he proves that a TA representation cannot be polynomially converted to an SA representation, which clearly shows that SA and CSAR are quite different concepts. His proof that not all planning instances have plans with an SA representation does thus not obviously carry over also to CSARs. Furthermore, he uses a planning language where actions are modelled as polynomial-size circuits. This coincides with our class FFP([**P**]). Hence, his hardness proofs are weaker than ours since we use the restricted STRIPS language in those cases. It should finally be noted that Liberatores Theorem 17 for the case of TA representations is a result similar to our Theorem 25, but we use different methods and different conditions.

Nebel (2000) defines a concept of compilation between planning languages. Although in some ways similar to our reformulation concept, there are also differences. A compilation is a function from a planning frame to another frame in a different planning language. This compilation need not be resource bounded but the resulting frame must be polynomially bounded in the original frame. The initial state and goal must then be possible to translate in polynomial time. That is, the first step corresponds to our function $r$ while the second step essentially corresponds to our function $R$. However, Nebel only considers compilation between planning languages and also requires a concept of modularity that is not present in our approach. Furthermore, his focus is not on the complexity of the decision problem but on the question whether the size of solutions is preserved by compilations.

## 7.5 Conclusions and Open Questions

The current status of our knowledge about non-uniform compact representations can be visualized as in Figure 6. The outer box represents the set of all solvable STRIPS instances while the inner boxes represent the subsets where at least one plan for each instance has a CSAR, CRAR or polynomial macro plan. We know classes where at least one plan for each instance is guaranteed to have a polynomial macro plan, like Towers of Hanoi and 3S. We also know classes where at least one plan for each instance has a CRAR but we do not know if those plans also have a polynomial macro plan. Construction 16 is such an example. It is an open question if a plan can have a CRAR but no polynomial macro plan. We further know classes where each instance has a plan with a CSAR but where we do not know if they also have a CRAR, for example, the class of reversible systems. However, Construction 26 is a class where all instances have a plan with a CSAR but no plan with a CRAR, so this case provides a strict separation between the CSAR and CRAR concepts. Whether there are classes of STRIPS instances where no plan has a CSAR remains an open question, though.

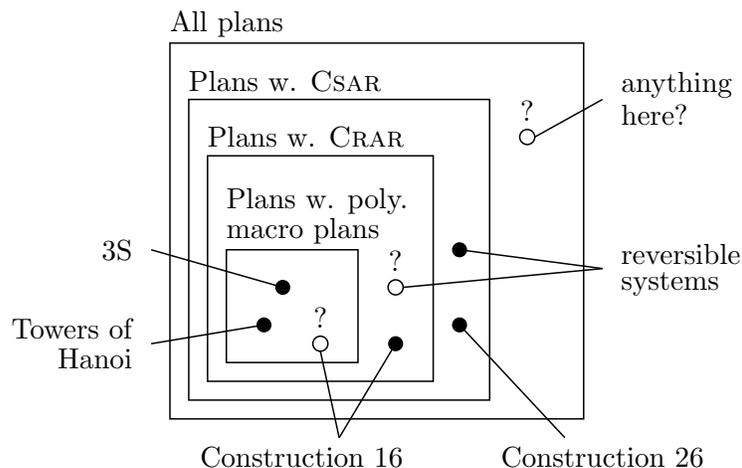

Figure 6: Current status for CSARS and CRARS.

We thus have the following chain of inclusions

$$\text{polynomial macro plans} \subseteq \text{CRAR} \subset \text{CSAR} \subseteq \text{STRIPS},$$

where we do not know if the first and last inclusions are strict.

Theorem 15 may seem weak since it is only conditioned with $\mathbf{P} \neq \mathbf{NP}$. Using similar techniques as in the proofs of Lemma 27 and Corollary 37 we could encode $\mathbf{QBF}$ with arbitrary number of alternating quantifiers and, hence, push the result up in the polynomial hierarchy. However, it remains an open question if the condition could be strengthened all the way to $\mathbf{P} \neq \mathbf{PSPACE}$.

While we have argued that a number of results hold also when restricted to unary instances, and in some cases also other restrictions, this is otherwise a largely unexplored area. Little is currently known about how various structural and other restrictions affect the results in this paper. This applies both to whether plans have CSARS and CRARS and to whether they have polynomial-size macro plans.

Just as we consider the non-uniform case of compact representations of single plans for single instances, it might also be interesting to consider the non-uniform case of reformulation and of adding information. However, this seems not straightforward since we could always reformulate an instance to a single bit telling whether the instance is solvable or not. Such a reformulation is clearly not interesting so additional criteria are necessary.

We have previously (Bäckström & Jonsson, 2011a) defined a complexity measure based on padding which is intended to be insensitive to plan length. This concept seems related to Nebel's compilations, although the two concepts are not identical or directly comparable. It is thus reasonable to believe that also compact representations and padded complexity are somehow related, especially since padded complexity was motivated by instances with long plans. However, we do not yet know what this relationship is. Furthermore, it would be interesting to consider compilations where we look at the size of compact representations of plan rather than the size of explicit plans.

## Acknowledgments

Malte Helmert, Anders Jonsson and the anonymous reviewers of this paper and the earlier conference version have provided valuable comments and suggestions.